\documentclass{article}

\usepackage{microtype}
\usepackage{graphicx}
\usepackage{subfigure}
\usepackage{booktabs} 

\usepackage{hyperref}

\usepackage[accepted]{icml2024}

\usepackage{amsmath}
\usepackage{amssymb}
\usepackage{mathtools}
\usepackage{amsthm}
\usepackage{graphicx}
\usepackage{wrapfig}
\usepackage{multirow}
\usepackage{color, colortbl}
\definecolor{LightCyan}{rgb}{0.88,1,1}
\definecolor{LightGreen}{rgb}{0.56, 0.93, 0.56}
\newcommand{\CC}[1]{\cellcolor{LightGreen}}
\usepackage{amsmath}
\usepackage{enumitem}

\newcommand{\score}{SCoRe}

\usepackage[capitalize,noabbrev]{cleveref}

\theoremstyle{plain}
\newtheorem{theorem}{Theorem}[section]

\theoremstyle{definition}

\theoremstyle{remark}

\usepackage[disable,textsize=tiny]{todonotes}

\icmltitlerunning{Submodular Combinatorial Representation Learning}

\begin{document}

\twocolumn[
\icmltitle{SCoRe: Submodular Combinatorial Representation Learning}



\icmlsetsymbol{equal}{*}

\begin{icmlauthorlist}
\icmlauthor{Anay Majee}{utd}
\icmlauthor{Suraj Kothawade}{google}
\icmlauthor{Krishnateja Killamsetty}{ibm}
\icmlauthor{Rishabh Iyer}{utd}

\end{icmlauthorlist}

\icmlaffiliation{utd}{Department of Computer Science, The University of Texas at Dallas, Richardson, TX, USA}
\icmlaffiliation{google}{Google Research, Sunnyvale, CA, USA}
\icmlaffiliation{ibm}{IBM Research, San Jose, CA, USA}

\icmlcorrespondingauthor{Anay Majee}{anay.majee@utdallas.edu}
\icmlcorrespondingauthor{Rishabh Iyer}{rishabh.iyer@utdallas.edu}

\icmlkeywords{Submodular Functions , Contrastive Learning , Supervised Learning , Representation Learning}

\vskip 0.3in
]



\printAffiliationsAndNotice{}  

\begin{abstract}
   In this paper we introduce the \textbf{SCoRe}\footnote{Project page:~\url{https://anaymajee.me/assets/project_pages/score.html}.} (\textbf{S}ubmodular \textbf{Co}mbinatorial \textbf{Re}presentation Learning) framework, a novel approach in  representation learning that addresses inter-class bias and intra-class variance. 
\score\ provides a new combinatorial viewpoint to representation learning, by introducing a family of loss functions based on set-based submodular information measures.
We develop two novel combinatorial formulations for loss functions, using the \emph{Total Information} and \emph{Total Correlation}, that naturally minimize intra-class variance and inter-class bias.
Several commonly used metric/contrastive learning loss functions like supervised contrastive loss, orthogonal projection loss, and N-pairs loss, are all instances of \score, thereby underlining the versatility and applicability of \score\ in a broad spectrum of learning scenarios.
Novel objectives in \score\ naturally model class-imbalance with up to 7.6\% improvement in classification on CIFAR-10-LT, CIFAR-100-LT, MedMNIST, 2.1\% on ImageNet-LT, and 19.4\% in object detection on IDD and LVIS (v1.0), demonstrating its effectiveness over existing approaches.\looseness-1

\end{abstract}

\section{Introduction}
\label{sec:intro}
\begin{figure}[t]
  \begin{center}
    \includegraphics[width=0.9\columnwidth]{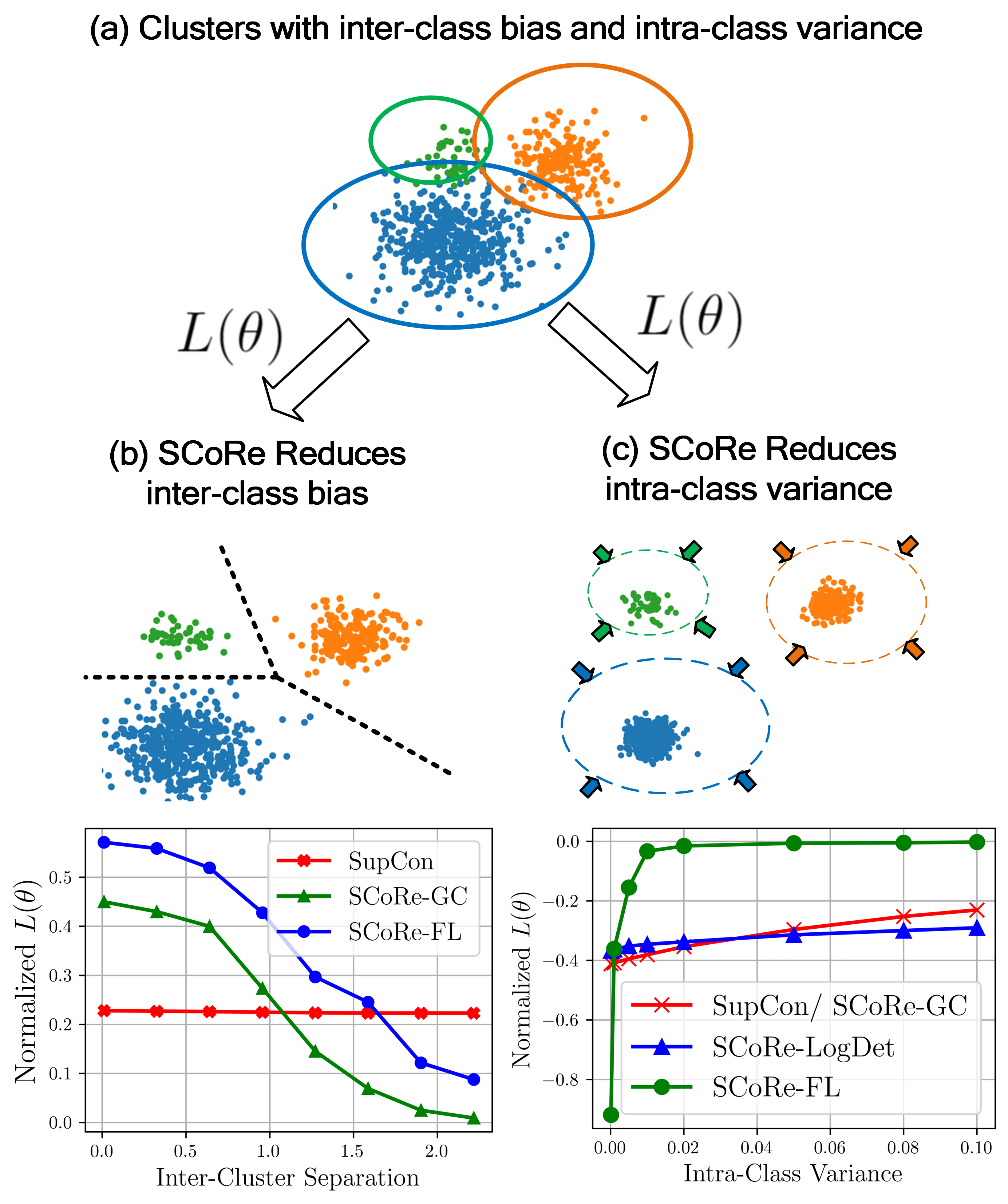}
  \end{center}
  \caption{\textbf{Objectives in \score\ are resilient to inter-class bias and intra-class variance in long-tail settings}. Applying $L(\theta)$ to (a) reduces inter-class bias by promoting inter-cluster separation in (b) while reducing intra-class variance in (c) by inducing intra-cluster compactness.\looseness-1}
  \vspace{-2ex}
  \label{fig:title_fig}
\end{figure}
Visual Object Recognition in real-world scenarios prominently features a long-tail distribution, where abundant (head) and rare (tail) objects coexist. 
The Open Long-Tail Recognition (OLTR) benchmark, as introduced in~\cite{oltr}, tackles the dual challenges of imbalanced and few-shot learning within a single, streamlined training framework. 
Notably, robust recognition necessitates that head and tail classes share visual features to compensate for the sparse examples in tail classes, a concept underscored in~\cite{oltr, liu2022oltr}. 
However, this sharing can inadvertently lead to confusion and erroneous predictions between visually similar objects also known as \textbf{inter-class bias}~\cite{agcm, metadet}, depicted in \cref{fig:title_fig}. 
Moreover, the prevalence of head classes often biases models~\cite{resnet, vgg} towards them, adversely affecting performance on tail classes. 
The considerable variability within head classes, have been shown to generate local sub-centers~\cite{Deng2020SubcenterAB} that intensify this bias and contribute to substantial \textbf{intra-class variance}, also showcased in \cref{fig:title_fig}.
To effectively navigate these complexities (further detailed in \cref{sec:rel_work}), \emph{the underlying model must adeptly minimize both inter-class bias and intra-class variance}. We highlight in \cref{sec:rel_work} that existing approaches adopt metric/contrastive learners like SupCon~\cite{supcon2020} to overcome the aforementioned challenges. Although State-of-the-Art (SoTA) approach SupCon shows appreciable variation to intra-class variance, as shown in \cref{fig:title_fig}, it fails to apply a significant penalty to inter-class bias. This highlights the need for a joint objective that can penalize both pitfalls during model training.\looseness-1 

\begin{figure}[t]
        \centering
        \includegraphics[width=0.9\columnwidth]{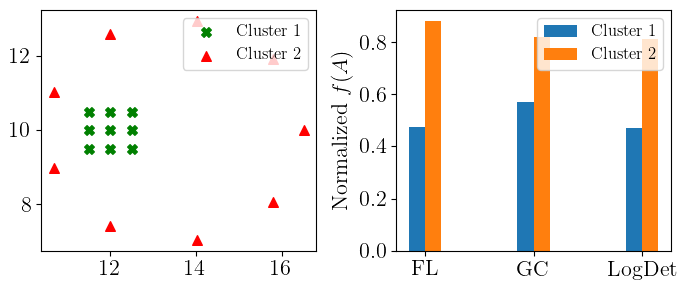}
        \vspace{-2ex}
        \caption{\small \textbf{Submodular functions} $f(A)$ in \score\ model diversity (A = Cluster 1) and cooperation (A = Cluster 2).\looseness-1}
        \label{fig:submodular_function}
        \vspace{-2ex}
\end{figure}

\begin{figure}[t]
  \begin{center}
    \includegraphics[width=0.9\columnwidth]{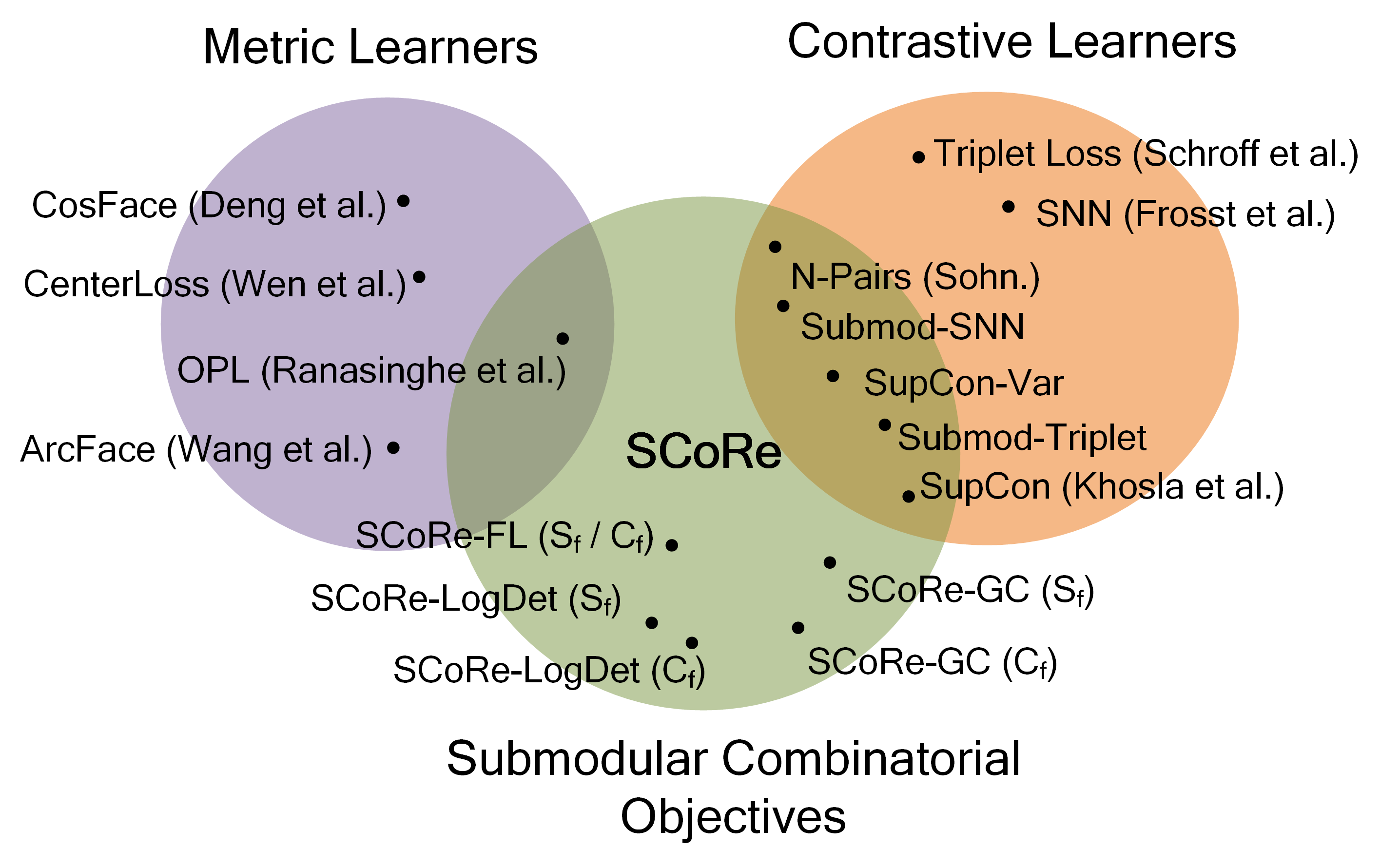}
  \end{center}
  \caption{\textbf{Overview of Combinatorial Objectives} in \score\ with respect to contrastive and metric learners.}
  \vspace{-3ex}
  \label{fig:overview}
\end{figure}

To address these limitations, we introduce a \textbf{S}ubmodular \textbf{Co}mbinatorial \textbf{Re}presentation Learning (\textbf{\score}) framework presenting a \textbf{family of combinatorial loss functions, as outlined in \cref{tab:comb_overview}, designed to effectively address the dual challenges of intra-class variance and inter-class bias in long-tail recognition}.
Our method leverages a combinatorial perspective by formulating the input dataset as a collection of sets (see \cref{sec:method}), facilitating the use of set-based functions~\cite{fujishige} as potent learning objectives.
Specifically, \score\ utilizes submodular functions, which model cooperation~\cite{submod_cooperation} (similarity) when minimized, and diversity~\cite{submod_diversity, Kulesza_2012} (dis-similarity) when maximized, due to their property of diminishing marginal returns~\cite{fujishige}.
This is illustrated in \cref{fig:submodular_function} where a submodular function $f(A)$ over a set $A$ has a low value in the presence of low intra-class variance (cluster 1) modelling \emph{cooperation} and has a high value otherwise (cluster 2) modeling \emph{diversity}.
We capitalize on these intrinsic properties of submodular functions to design a family of objectives as shown in \cref{tab:comb_overview} based on two well-known formulations- \emph{Total Information} and \emph{Total Correlation} which model total feature information and information gain on adding novel instances in a class respectively.
Instances in \score\ are driven by the strategic choice of submodular combinatorial functions as discussed in \cref{sec:submod_obj} where some objectives inherently demonstrate class-balancing (SCoRe-FL) while jointly modelling inter-class bias and intra-class variance.
We demonstrate this in \cref{fig:title_fig}(b, c) where \score\ objectives exhibit larger relative variations in tackling both intra-class variance and inter-class bias compared to state-of-the-art approaches, making them superior for model training.
The primary contributions of this paper are as follows:\looseness-1
\vspace{-2ex}
\begin{itemize}[leftmargin=*]
    \item We introduce the novel \score\ framework, introducing a \textbf{set-based combinatorial viewpoint} to representation learning under long-tail settings.\looseness-1
    \item We show that \textbf{\score\ generalizes several existing metric/contrastive learning approaches} like SupCon, N-pairs and OPL losses which are instances of \score\ as shown in \cref{fig:overview} exhibiting combinatorial properties.\looseness-1
    \item Objectives in \score\ \textbf{learn better features within fewer number of training epochs} in the \emph{imbalanced} (\cref{fig:score_convergence}) setting while \textbf{demonstrating faster convergence} in the \emph{balanced} setting as noted in \cref{app:abl_convergence}).\looseness-1
    \item Lastly, introduction of \score\ objective functions result in outperforming State-of-the-Art (SoTA) approaches by \textbf{up to 7.6\% for classification} tasks and \textbf{19.4\% for object detection} tasks for class-imbalanced settings like CIFAR-10-LT, CIFAR-100-LT, MedMNIST, Imagenet-LT, IDD and LVIS (v1.0).\looseness-1
\end{itemize}
\vspace{-2ex}

\section{Related Work}
\label{sec:rel_work}
\paragraph{Long-tail Learning:} Visual recognition tasks in long-tail learning, which focus on learning from few over-represented and many under-represented classes, have traditionally tackled class imbalance in datasets through over-sampling of rare classes~\citep{aug_smote} or under-sampling of abundant ones~\citep{aug_cb,aug_dist_align}. However, these methods alter the original dataset's distribution, hindering model generalization. An alternative method is re-weighting class probabilities during learning, either by giving more weight to tail classes or restricting gradient updates for abundant classes~\citep{reweight_nips19,reweight_nips17,bbn,reweight_cb_20}. Despite these efforts, recent research~\cite{bbn} suggests that re-weighting strategies lead to sub-optimal representation learning. Progress has been made through metric/contrastive learning strategies~\citep{paco,gpaco,cont_cvpr21,bcl} and two-stage training objectives requiring extensive negative label information~\citep{supcon2020, simclr2020} which are challenging to implement on large datasets like ImageNet-LT~\cite{oltr}. Vision transformers~\cite{vit}, combined with other techniques~\citep{vit_eccv_22,vit_cvpr23}, have shown significant improvements, albeit at a high computational cost. Recent SoTA approaches, such as~\cite{paco, gpaco, glmc, proco}, blend data augmentation with contrastive learning. GPaCo innovates with parametric learnable class centers~\cite{mocov2}, while GLMC~\citep{glmc} introduces a loss combining global MixUp~\citep{mixup}, local CutMix~\citep{cutmix}, and a cumulative soft label reweighted loss. Notably, all SoTA methods employ some form of contrastive learning, underscoring its importance in long-tail learning.\looseness-1
\vspace{-2ex}

\paragraph{Metric and Contrastive Learning:} In supervised learning, traditional models using Cross-Entropy (CE) loss~\citep{celoss} struggle with class imbalance and noisy labels. Metric learning approaches~\citep{arcface, cosface, opl, msloss} address this by learning distance~\citep{triplet} or similarity~\citep{arcface, cosface} metrics, promoting orthogonality in feature space~\citep{opl} and enhancing class-specific feature discrimination. Contrastive learning, derived from noise contrastive estimation~\citep{nce}, is prevalent in self-supervised learning~\citep{simclr2020, moco, mocov2} where label information is absent during training. In supervised domains, SupCon~\citep{supcon2020} focuses on learning feature clusters, not just aligning features to centroids. Triplet loss~\citep{triplet} contrasts one positive and negative pair, while N-pairs~\citep{n_pairs} loss uses multiple negative pairs, and SupCon uses multiple positive and negative pairs. Lifted-Structure loss~\citep{lsl} contrasts positives with the hardest negatives. SupCon is similar to Soft-Nearest Neighbors loss~\citep{snn}, maximizing class entanglements. Despite successes, these methods rely on pairwise similarity metrics and may not ensure disjoint cluster formation.\looseness-1
\vspace{-2ex}

\paragraph{Submodular Functions} are set functions that satisfy a natural diminishing returns property. A set function $f: 2^\mathcal{V} \rightarrow \mathbb{R}$ (on a ground-set $\mathcal{V}$) is submodular if it satisfies $f(X) + f(Y) \geq f(X \cup Y)  + f(X \cap Y), \forall X, Y \subseteq \mathcal{V}$~\citep{fujishige}. 
These functions have been studied extensively in the context of data subset selection~\citep{killamsetty_automata, prism, jain2023efficient}, active learning~\citep{wei15_subset, talisman, Beck2021EffectiveEO, Kaushal_2019} and video-summarization~\citep{vid_sum_2019,kaushal2019framework,kaushal2021vid_summary}.
Submodular functions are capable of modelling diversity, relevance, set-cover etc. which allows them to discriminate between different classes or slices of data while ensuring the preservation of most relevant features in each set.
Very recent developments in the field have applied submodular functions like Facility-Location in metric learning~\citep{fl_repr_learning}.
Minimizing submodular functions $f(A)$ over a set $A$ models cooperation~\cite{submod_cooperation} between samples while maximizing submodular functions model diversity~\cite{submod_diversity} making these functions suitable for learning diverse feature clusters in representation learning tasks which is yet to be studied in literature.\looseness-1

To the best of our knowledge , we are the first to demonstrate that combinatorial objectives using submodular functions are superior in creating tighter and well-separated feature clusters for representation learning. 
We are fore-runners in showing through \cref{sec:experiments} that \score\ generalizes several existing contrastive learning approaches while some others can be reformulated as a superior submodular variant.\looseness-1
\vspace{-2ex}

\section{\score: Submodular Combinatorial Representation Learning Framework}
\label{sec:method}

Supervised training for representation learning tasks proceed with learning a \textbf{feature extractor} $F(I, \theta)$~\citep{alexnet, vgg, resnet} which projects an input image $I$ into a $D$ dimensional feature space $r$, where $r = F(I, \theta) \in \mathbb{R}^{D}$, given parameters $\theta$.
A \textbf{classifier} $Clf(r, \theta)$ operates on the embeddings produced by $F$ to categorize an input image $I$ in the input dataset $\mathcal{T}$ into its corresponding class label $c_{i}$, where $i \in [1,2,...C]$. 
From our discussions in \cref{sec:intro}, representation learning in the long-tail setting requires the learning of generalizable features in $F(I, \theta)$, minimizing the impact of inter-class bias and intra-class variance.
Previous literature in \cref{sec:rel_work} has shown promise in this directions by promoting the learning of discriminative features by $F(I, \theta)$, attributed to be a function of a learning objective $L(\theta)$.

\begin{table*}[ht]
      \caption{\textbf{Summary of various instantiations of \score\ } and their respective combinatorial properties (detailed derivations in section \ref{app:proof_submod} of the appendix).}
      \centering
      \small
      \setlength{\tabcolsep}{1pt}
      \resizebox{0.75\textwidth}{!}{\begin{tabular}{l|c|c}
            \hline
            \multirow{ 2}{*}{\textbf{Objective Function}}  & 
            \multirow{ 2}{*}{\textbf{Equation $L(\theta)$}}& 
            \textbf{Combinatorial} \\
             & & \textbf{Property}\\
            \hline \hline
            Triplet Loss \cite{triplet} & 
            $ \sum_{k = 1}^{|C|}\frac{1}{|A_k|}[ \sum_{\substack{i,p \in A_{k}, \\ n \in \mathcal{V} \setminus A_{k}}} \max (0, D_{ip}^2(\theta) - D_{in}^2(\theta) + \epsilon)]$ &
            Not Submodular \\

            & & \\

            SNN \cite{snn} &
            $\sum_{k = 1}^{|C|}\frac{-1}{|A_k|}\sum_{i \in A_{k}} {\log \sum_{j \in A_{k}} \exp(S_{ij}(\theta)) + \frac{1}{|A_k|} \log\sum_{\substack{i \in A_{k} \\ j \in \mathcal{V} \setminus A_{k}}} \exp(S_{ij}(\theta))}$ &
            Not Submodular \\

            & & \\ \hline

            N-Pairs Loss \cite{n_pairs} &
            $\sum_{k = 1}^{|C|} \frac{-1}{|A_k|}{\sum_{i,j \in A_{k}} S_{ij}(\theta) + \frac{1}{|A_k|}\sum_{i \in A_{k}} log(\sum_{j \in \mathcal{V}} S_{ij}(\theta) - 1)} $ &
            Submodular \\
            
            & & \\

            OPL \cite{opl} &
            $\sum_{k = 1}^{|C|} \frac{1}{|A_k|}( 1 - \sum_{i,j \in A_{k}} S_{ij}(\theta)) + \frac{1}{|A_k|}\sum_{i \in A_{k}} \sum_{j \in \mathcal{V} \setminus A_{k}} S_{ij}(\theta)$ &
            Submodular \\

            & & \\ 
            
            SupCon \cite{supcon2020} &
            $\sum_{k = 1}^{|C|} [\frac{-1}{|A_{k}|} \sum_{i,j \in A_{k}} S_{ij}(\theta) ] + \sum_{i \in A_{k}}\frac{1}{|A_{k}|}\log(\sum_{j \in V}exp(S_{ij}(\theta)) - 1)$ &
            Submodular \\
            
            & & \\ \hline

            Submod-Triplet &
            $\sum_{k = 1}^{|C|} \frac{1}{|A_k|}\sum_{\substack{i \in A_{k} \\ n \in \mathcal{V} \setminus A_{k}}} S_{in}^{2}(\theta) - \sum_{i,p \in A} S_{ip}^{2}(\theta)$ &
            Submodular \\

            & & \\

            Submod-SNN &
            $\sum_{k = 1}^{|C|} \frac{1}{|A_k|}\sum_{i \in A_{k}} [\log \sum_{j \in A_{k}} \exp(D_{ij}(\theta)) + \log\sum_{j \in \mathcal{V} \setminus A_{k}} \exp(S_{ij}(\theta))]$ &
            Submodular \\

            & & \\

            SupCon-Var &
            $\sum_{k = 1}^{|C|} \frac{-1}{|A_k|}{\sum_{i,j \in A_{k}} S_{ij}(\theta) + \frac{1}{|A_k|}\sum_{i \in A_{k}} log \sum_{j \in \mathcal{V} \setminus A_{k}} \exp(S_{ij}(\theta))}$ &
            Submodular \\

            & & \\ \hline

            \score-GC [$S_f$] (ours) &
            $\sum_{k = 1}^{|C|} \frac{1}{|A_k|}[\sum_{i \in A_{k}}\sum_{j \in \mathcal{V} \setminus A_{k}}S_{ij}(\theta) - \lambda \sum_{i, j \in A_{k}} S_{ij}(\theta)]$ &
            Submodular \\

            & & \\

            \score-GC [$C_f$] (ours) &
            $\sum_{k = 1}^{|C|} \frac{\lambda}{|A_k|}\sum_{i \in A_{k}}\sum_{j \in \mathcal{V} \setminus A_{k}}S_{ij}(\theta) $&
            Submodular \\

            & & \\

            \score-LogDet [$S_f$] (ours) &
            $\sum_{k = 1}^{|C|} \frac{1}{|A_k|}\log \det (S_{A_k}(\theta) + \lambda \mathbb{I}_{|A_k|})$ &
            Submodular \\ 

            & & \\

            \score-LogDet [$C_f$] (ours) &
            $\sum_{k = 1}^{|C|} \frac{1}{|A_k|}[\log \det (S_{A_k}(\theta) + \lambda \mathbb{I}_{|A_k|}) - \log \det (S_{\mathcal{V}}(\theta) + \lambda \mathbb{I}_{|\mathcal{V}|})]$ &
            Submodular \\ 

            & & \\
            
            \score-FL [$C_{f}$/ $S_{f}$] (ours) &
            $\sum_{k = 1}^{|C|} \frac{1}{|\mathcal{V}|}\sum_{i \in \mathcal{V} \setminus A_{k}} max_{j \in A_{k}} S_{ij}(\theta)$ &
            Submodular \\            
            \hline
      \end{tabular}}\\
      \label{tab:comb_overview}
      \vspace{-2ex}
\end{table*}

\score\ introduces a family of \textbf{Combinatorial Loss Functions}, $L(\theta)$ as shown in \cref{tab:comb_overview}, which trains the feature extractor $F$ over all classes $C$ in the dataset $\mathcal{T}$. 
\score\ differs from existing approaches in the field through the introduction of a \emph{set-based combinatorial viewpoint} by defining a dataset $\mathcal{T}$ as a collection of sets, $\mathcal{T} = \{A_1,A_2, \cdots, A_{|C|}\}$ over classes (now represented as sets $A_k$) in $\mathcal{T}$.
Objectives in \score\ further differ from previous works, by adopting submodular functions~\cite{fujishige, alt_gen_submod} as learning objectives, which inherently model \emph{cooperation}~\cite{submod_cooperation} and \emph{diversity}~\cite{submod_diversity} by the virtue of their diminishing marginal returns property~\cite{fujishige}.
Adopting such a formulation in representation learning thus enforces cooperation within each class while promoting diversity between classes through $L(\theta)$, leading to the learning of discriminative class-specific features.
Additionally, submodular functions \textbf{model the information contained in a set $A$ (class in \score) irrespective of its size}, motivating their application in long-tail settings.
We adapt this viewpoint in \cref{sec:submod_obj} and propose two distinct formulations of combinatorial objectives - \emph{Total Information} ($L_{S_f}(\theta)$) and \emph{Total Correlation} ($L_{C_f}(\theta)$), that inherently model inter-class bias and intra-class variance to learn compact yet well-separated class-specific feature clusters.\looseness-1

Training and evaluation of models through the \score\ framework follows~\citet{supcon2020} and occurs in two stages. 
The first stage trains $F(I, \theta)$ to learn discriminative features through the newly introduced objectives $L(\theta)$, while the second stage trains the classifier $Clf(F(I,\theta))$ ($F$ is frozen) using the standard cross-entropy loss~\cite{celoss}.\looseness-1

\subsection{Combinatorial Loss Functions}
\label{sec:submod_obj}
Given an input data batch (referred to as ground set), $\mathcal{V} = \overset{|C|}{\underset{k=1}{\cup}} A_k$, and a submodular function $f(A_k; \theta)$ over a set $A_k$, we define a loss $L(\theta)$ which is an \emph{instantiation of submodular information functions}.  
Here, $A_k$ is a set representing each class in the dataset $\mathcal{T}$, $k \in [1,C]$ and $f$ is defined with similarity kernels $S$, which depends on the parameters $\theta$.
We propose two flavors of information measures from~\cite{alt_gen_submod}, the Total Submodular Information: $S_f(A_1, A_2, A_3, \dots, A_{|C|}) = \overset{|C|}{\underset{k=1}{\sum}}f(A_k)$ and the Total Submodular Correlation: $C_f(A_1, A_2, A_3, \dots, A_{|C|}) = \sum_{k=1}^{|C|} f(A_k) - f(\overset{|C|}{\underset{k=1}{\cup}}A_k)$. 
In context of representation learning, $S_f$ captures the total information contained in an object class (referred to as set) $A_k \in \mathcal{T}$ while $C_f$ captures the gain in information when new features are added to the set $A_k$.
Using the aforementioned $S_f$ and $C_f$ formulations, we can define two variants of combinatorial loss functions $L(\theta)$ for long-tail recognition tasks:
\vspace{-2ex}
\begin{align}
\label{eq:score_formulation}
\begin{split}
   L_{S_f}(\theta) =& \overset{|C|}{\underset{k=1}{\sum}} \frac{1}{N_f(A_k)}\,\,f(A_k; \theta), \\
   L_{C_f}(\theta) =& \overset{|C|}{\underset{k=1}{\sum}} \frac{1}{N_f(A_k)}\,\, \Biggl[f(A_k; \theta) - f(\overset{|C|}{\underset{k=1}{\cup}}A_k; \theta)\Biggr]
\end{split}
\end{align}
Where, $N_f(A_k)$ is the normalization constant over each set in $\mathcal{T}$.
As discussed in \cref{sec:method}, minimizing a submodular function $f(A_k; \theta)$ through $L(\theta)$ over a set $A_k$ captures \emph{cooperation}~\cite{submod_cooperation,iyer2015polyhedral} between samples in the set, while maximizing $f(A_k; \theta)$ captures diversity / coverage~\cite{submod_diversity,iyer2015submodular} between sets.
Consequently, objective \textbf{$L_{S_f}(\theta)$ which minimizes $S_f$ enforces intra-cluster compactness} by maximizing cooperation within each set $A_k$ (minimizing $f(A_k; \theta)$ over each set $A_k$).
Further, \textbf{$L_{C_f}(\theta)$ which minimizes $C_f$ enforces both intra-cluster similarity (first term in $L_{C_f}$) and inter-cluster separation} (by maximizing $f(\cup_k A_k)$). $L_{C_f}$ achieves inter-cluster separation by maximizing \emph{diversity} between orthogonal sets in $\mathcal{T}$.
We introduce several instantiations of the above formulations in our framework \emph{based on the choice of the underlying submodular function $f(A_k; \theta)$} producing a family of objective functions as shown in Table \ref{tab:comb_overview} for representation learning tasks.\looseness-1

\subsubsection{Instantiations of Combinatorial Objectives in \score}
\label{sec:score}
By varying the choice of submodular function $f(A)$ in \score\ we propose several novel objective functions in \cref{tab:comb_overview}. 
It is interesting to note that, our combinatorial objectives adopt a pairwise similarity kernel $S_{ij}(\theta)$ similar to existing approaches discussed in \cref{sec:rel_work}. 
However, \score\ objectives utilize the similarity kernel only to compute \emph{feature interactions between samples}, \textbf{differing from existing approaches in the aggregation of pairwise similarities to compute total information/ correlation} for a class $A_k \in \mathcal{T}$.
In practice, we adopt the cosine similarity metric $S_{ij}(\theta)$ as used in~\citet{supcon2020}, defined as $S_{ij}(\theta) = \frac{F(I_{i}, \theta)^{\text{T}} \cdot F(I_{j}, \theta)}{||F(I_{i}, \theta)|| \cdot ||F(I_{j}, \theta)||}$ to produce three novel instantiations namely - \score-FL, \score-GC and \score-LogDet based upon popular submodular functions in~\citet{iyer2015submodular,vid_sum_2019,kaushal2019framework} - Graph-Cut, Facility-Location and Log-Determinant respectively.\looseness-1

\textbf{\score-FL} based objective function minimizes the maximum similarity $S_{ij}$ (where $i \neq j$) between orthogonal sets (different class labels).\looseness-1
\vspace{-2ex}
\begin{theorem}
\label{thm:fl}
    If $f(A, \theta) = {\underset{i \in \mathcal{V}}{\sum}} {\underset{j \in A}{\max}} S_{ij}(\theta)$ represents the facility-location function over a set $A$ then, $L_{S_f}(\theta)$ and $L_{C_f}(\theta)$ shown in \cref{eq:fl} represents the \score-FL objective with $N_f(A_k) = |\mathcal{V}|$. Both $L_{S_f}(\theta)$ and $L_{C_f}(\theta)$ differ by a constant.\looseness-1
\end{theorem}
\vspace{-3ex}
\begin{align}
\begin{split}
L_{S_f}(\theta) =& \overset{|C|}{\underset{k=1}{\sum}}\frac{1}{|\mathcal{V}|}\sum_{i \in \mathcal{V} \backslash A_k} {\underset{j \in A_k}{\max}} S_{ij}(\theta) + 1, \\
L_{C_f}(\theta) =& \overset{|C|}{\underset{k=1}{\sum}}\frac{1}{|\mathcal{V}|}\sum_{i \in \mathcal{V} \backslash A_k} {\underset{j \in A_k}{\max}} S_{ij}(\theta)   
\end{split}
\label{eq:fl}
\end{align}
Evident in its form in \cref{eq:fl}, \score-FL promotes large inter-cluster separation by minimizing the similarity between the hardest negative pair between $\mathcal{V} \setminus A_k$ and $A_k$.Additionally, \textbf{\score-FL inherently introduces a class balancing} property making it an inevitable choice for learning in long-tail settings detailed in \cref{sec:properties}.\looseness-1

\textbf{\score-GC} objective minimizes the feature similarity between representations of a positive set $A_{k}$ and the remaining negative sets $\mathcal{V} \setminus A_{k}$ while maximizing the similarity among features in each set $A_{k}$.\looseness-1
\begin{theorem}
\label{thm:gc}
    If $f(A, \theta) = {\underset{i \in A, j\in V}{\sum}} S_{ij}(\theta) - \lambda \sum_{i, j \in A} S_{ij}(\theta)$ represents the Graph-Cut function over a set $A$ then, $L_{S_f}(\theta)$ and $L_{C_f}(\theta)$ shown in \cref{eq:gc} represents the \score-GC objective, normalized by $|A_k|$.\looseness-1
\end{theorem}
\vspace{-4ex}
\begin{align}
\begin{split}
L_{S_f}(\theta) =& \overset{|C|}{\underset{k=1}{\sum}}\frac{1}{|A_k|}\Biggl[{\underset{\substack{i \in A_{k}, \\j\in V \setminus A_k}}{\sum}} S_{ij}(\theta) - \lambda \sum_{i, j \in A_{k}} S_{ij}(\theta)\Biggr], \\
L_{C_f}(\theta) =& \overset{|C|}{\underset{k=1}{\sum}} \frac{\lambda}{|A_k|} {\underset{\substack{i \in A_{k},\\ j\in V \setminus A_k}}{\sum}} S_{ij}(\theta) 
\end{split}
\label{eq:gc}
\end{align}
The hyper-parameter $\lambda$~\citet{submod_diversity} controls the weightage of the loss to intra-class compactness over inter-cluster separation and is ablated upon in \cref{tab:abl_lambda}. It is interesting to note that \textbf{\score-GC jointly models inter-cluster separation and intra-cluster compactness}. Minimizing the first term in \score-GC ($-\sum_{i, j \in A_k} S_{ij}$) promotes intra-cluster compactness, while the second term ($\sum_{i \in A_k}\sum_{j \in \mathcal{V} \setminus A_k} S_{ij}(\theta)$) penalizes cluster overlaps i.e. promoting inter-class separation.
Note that Orthogonal Projection Loss (OPL) and a version of Triplet Loss are special cases of of the GC based loss function.\looseness-1

\textbf{\score-LogDet} presents an unique perspective by modelling the volume of a set $A_k$ in the embedding space.
\begin{theorem}
\label{thm:logdet}
    If $f(A, \theta) = \log \det (S_{A}(\theta) + \lambda \mathbb{I}_{|A|})$ represents the Log-Determinant function over a set $A$ then, $L_{S_f}(\theta)$ and $L_{C_f}(\theta)$ shown in \cref{eq:logdet} represents the \score-LogDet objective which models the volume of each set $A_k \in \mathcal{T}$. $\mathbb{I}_{|A_k|}$ and $\mathbb{I}_{|\mathcal{V}|}$ indicate identity terms and $N_f(A_k) = |A_k|$, introduced for numerical stability.\looseness-1
\end{theorem}
\vspace{-4ex}
\begin{align}
\begin{split}
L_{S_f}(\theta) =& \overset{|C|}{\underset{k=1}{\sum}} \frac{1}{|A_k|}\log \det (S_{A_k}(\theta) + \lambda \mathbb{I}_{|A_k|}), \\
L_{C_f}(\theta) =& L_{S_f}(\theta) - \overset{|C|}{\underset{k=1}{\sum}}\frac{1}{|A_k|}\log \det (S_{\mathcal{V}}(\theta) + \lambda \mathbb{I}_{|\mathcal{V}|})   
\end{split}
\label{eq:logdet}
\end{align}

\begin{figure*}[t]
        \centering
        \includegraphics[width=0.8\textwidth]{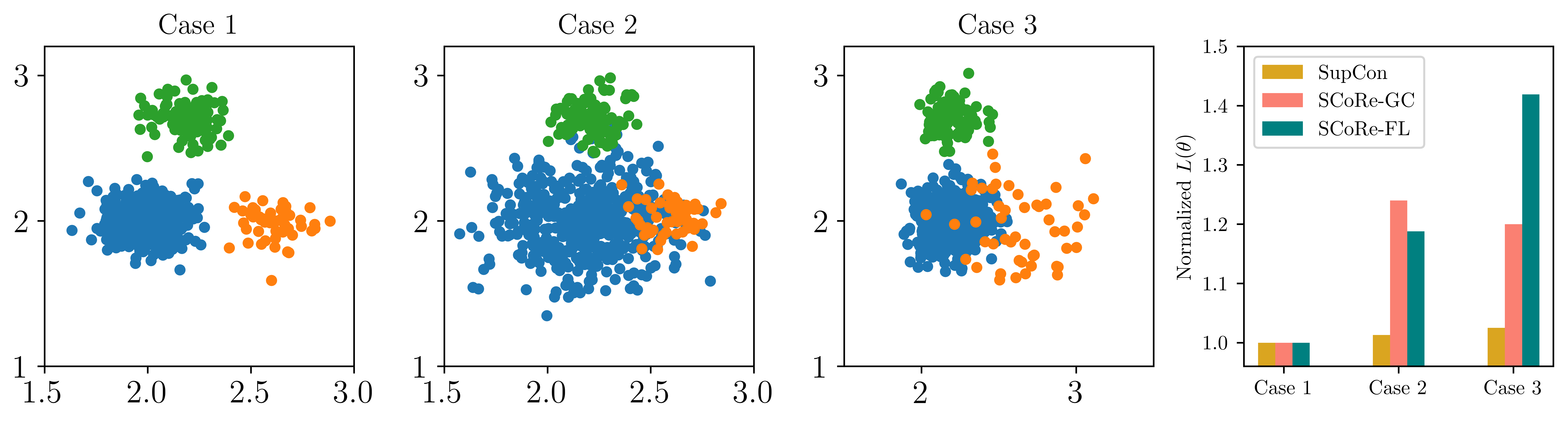}
        \vspace{-2ex}
        \caption{\textbf{Resilience to Intra-Class Variance and Inter-Class Bias under the Long-tail setting}. Case 1 demonstrates no intra-class variance and inter-class bias, while case 2 demonstrates larger variance for the head class wile case 3 demonstrates larger variance for the tail class inducing inter-cluster overlaps. The details of the experiment have been enclosed in \cref{app:exp_setup}. \looseness-1}
        \label{fig:synthetic_imbalanced}
        \vspace{-2ex}
\end{figure*}

Minimizing \score-LogDet over a set $A_k$ through $L_{S_f}$ \textbf{minimizes the cluster volume of $A_k$ inherently reducing intra-class variance}.
$L_{C_f}$ (which empirically we see works better) captures both intra-cluster similarity and inter-cluster dissimilarity by additionally maximizing the diversity in the feature space $\mathcal{V}$. 
Note that the additive term $\mathbb{I}_{|A_k|}$ indicates an identity term introduced for numerical stability.

Proofs for all theorems are provided in \cref{app:proof_submod_loss}. 
Our experiments in \cref{sec:experiments} indicate that adopting set-based objectives defined in \score\ outperforms existing metric/contrastive loss functions.\looseness-1 

\subsubsection{\score\ Generalizes Existing Metric/Contrastive Learning Objectives}
\label{sec:exiting_metric_learners}
From the formulations in \cref{tab:comb_overview} we observe that \textbf{\score\ generalizes to several metric/contrastive learning objectives which inherently demonstrate combinatorial properties}. 
We provide proofs in \cref{app:proof_submod}.
N-pairs~\cite{n_pairs} and OPL~\cite{opl} and SupCon~\cite{supcon2020} are all instances of \score\ and are submodular (in the $S_f$ form), exhibiting combinatorial properties.
On the other hand, Triplet~\cite{triplet} and SNN~\cite{snn} losses are not instances of \score\ in their original form while their algebraically modified forms - Submod-Triplet and Submod-SNN as shown in \cref{tab:comb_overview} are instances of \score\ and outperform their non-submodular counterparts.

Analyzing SoTA approach SupCon as shown in \cref{tab:comb_overview} we decompose it into an intra-class ($-\sum_{i, j \in A_k} S_{ij}$) and an inter-class ($\log(\sum_{i \in A_k, j \in V}exp(S_{ij})$) term. The \emph{inter-class term is computed over $\mathcal{V}$ which includes $A_k$, enforcing separation between samples in $A_k$ which is counter intuitive} to the task at hand, hindering its capability to overcome inter-class bias. This is demonstrated by the low variation in the SupCon objective in the presence of large inter-class bias in \cref{fig:title_fig}(b).
To overcome this pitfall in SupCon we introduce, SupCon-Var (row 9 of \cref{tab:comb_overview}) which minimizes the maximum similarity between a set $A_k$ and $\mathcal{V}\setminus A_k$ in the inter-class term emerging as a stronger objective over SupCon in overcoming inter-class bias (the intra-class term is similar to SupCon).\looseness-1

\subsubsection{Contrasting Instantiations of \score}
\label{sec:properties}
In this section we discuss some of the \emph{properties} of combinatorial objectives, through experiments on synthetic data (refer \cref{fig:synthetic_imbalanced}) which upholds our claim towards their application in long-tail recognition tasks.
Although their properties largely depends on the choice of the submodular function $f(A_i; \theta)$ we enlist some unique ones favorable to long-tail settings.\looseness-1
\vspace{-2ex}
\begin{itemize}[leftmargin=*]
    \item \textbf{\score\ Objectives model Information in a Set}. Unlike exiting approaches, the novel formulations in \score\ model total information ($L_{S_f}$) contained in a class ($A_k$) and the gain in information when new instances are added to $A_k$ ($L_{C_f}$), \emph{irrespective of the the size of the class}. This viewpoint in representation learning is novel to \score.\looseness-1
    \item \textbf{\score-FL demonstrates inherent Class-balancing}, equally weighting tail classes in contrast to head ones in model training. Unlike SoTA approaches~\cite{supcon2020,bcl, simclr2020} that scale linearly with the size of the set $|A_k|$, \score-FL has an inverse relation, and scales with respect to $\mathcal{V}\setminus A_k$ ($\sum_{i \in \mathcal{V} \setminus A_{k}} max_{j \in A_{k}} S_{ij}(\theta)$). This inherently introduces class balancing critical in long-tail settings. \cref{fig:synthetic_imbalanced} shows that, with increase in variance of the tail class (marked in \textcolor{orange}{orange}) over the head class in case 3, \score-FL and \score-GC provide a larger relative penalty to the model than SoTA contrastive learners~\cite{supcon2020, simclr2020}. Interestingly, the relative penalty applied by \score-FL is significantly larger in case 3 than in case 2, where the variance of the head class (marked in \textcolor{blue}{blue}) is larger than the tail. This highlights the class-balancing behavior in \score-FL, thereby establishing its effectiveness in long-tail recognition tasks.\looseness-1
    \vspace{-2ex}
    \item \textbf{\score-LogDet is a volumetric function} and models the volume of a feature cluster~\cite{fujishige}, minimizing which shrinks the cluster volume resulting in reduced intra-class variance ($L_{S_f}$). Additionally, maximizing the cluster volume over all sets ($L_{C_f}$) in the dataset maximizes diversity among clusters mitigating inter-class bias.\looseness-1
\end{itemize}


\section{Experiments}
\label{sec:experiments}
\label{res:classification}
We perform experiments on several long-tail vision benchmarks to show the effectiveness of objectives in \score.
\vspace{-3ex}
\subsection{Datasets and Experimental Setup}
\textbf{CIFAR-10-LT} consists of 10 disjoint classes with imbalance factors (IFs) ranging from 10 to 100~\cite{oltr}. The dataset is created by randomly sampling the balanced CIFAR-10~\cite{cifar} dataset based on an exponentially decaying function with IF as rate of decay.
Additionally, we introduce a pathological benchmark based on a \textit{step} distribution by exploiting the hierarchy already available (living vs non-living objects) in the dataset.
The data distributions of the adopted benchmarks are depicted in \cref{fig:data_dist_cifar}(a, b).\looseness-1

\begin{table*}[t]
    \centering
    \small
    \caption{\textbf{Multi-class classification performance (Top1-Accuracy \%) of combinatorial objectives in \score} (shaded in \colorbox{LightGreen}{Green}) against existing approaches in Longtail recognition for \textit{CIFAR-10-LT} and \textit{CIFAR-100-LT} datasets for varying Imbalance Factors (IF).\looseness-1}
    \label{tab:cifar_lt}
    \resizebox{0.8\textwidth}{!}{\begin{tabular}{c|l|lll|lll}
    \hline 
    \multirow{2}{*}{} & \multirow{2}{*}{ Method } & \multicolumn{3}{|c|}{ CIFAR-10-LT } & \multicolumn{3}{c}{ CIFAR-100-LT } \\
    \cline { 3 - 8 }
     & & IF=100 & 50 & 10 & 100 & 50 & 10 \\ \midrule \midrule
     & CE & 70.4 & 74.8 & 86.4 & 38.3 & 43.9 & 55.7 \\
     & Focal Loss \cite{focal_loss} & 70.38 & 76.72 & 86.66 & 38.41 & 44.32 & 55.78 \\
    \hline 
    \multirow{4}{*}{ Class/ Weight Balanced} & BBN \cite{bbn} & 79.82 & 82.18 & 88.32 & 42.56 & 47.02 & 59.12 \\
    & CB-Focal \cite{cb_focal} & 74.6 & 79.3 & 87.1 & 39.6 & 45.2 & 58 \\
    & LogitAjust \cite{logit_adj} & 80.92 & - & - & 42.01 & 47.03 & 57.74 \\
    & weight balancing \cite{weight_bal} & - & - & - & 53.35 & 57.71 & 68.67 \\ \hline 
    \multirow{3}{*}{ Augmentation Based} & Mixup \cite{mixup} & 73.06 & 77.82 & 87.1 & 39.54 & 54.99 & 58.02 \\
    & RISDA \cite{risda} & 79.89 & 79.89 & 79.89 & 50.16 & 53.84 & 62.38 \\
    & CMO \cite{cmo} & - & - & - & 47.2 & 51.7 & 58.4 \\ \hline
     \multirow{2}{*}{ Ensemble Classifier } & RIDE (3 experts) + CMO \cite{ride} & - & - & - & 50 & 53 & 60.2 \\
    & RIDE (3 experts) \cite{ride} & - & - & - & 48.6 & 51.4 & 59.8 \\ \hline
    \multirow{6}{*}{ SSL-Pretraining } & KCL \cite{kcl} & 77.6 & 81.7 & 88 & 42.8 & 46.3 & 57.6 \\
    & TSC \cite{tsc} & 79.7 & 82.9 & 88.7 & 42.8 & 46.3 & 57.6 \\
    & SSD \cite{ssd} & - & - & - & 46.0 & 50.5 & 62.3 \\ 
    & BCL \cite{bcl} & 84.32 & 87.24 & 91.12 & 51.93 & 56.59 & 64.87 \\ \cline {2 - 8 }
    & PaCo \cite{paco} & 85.11 & 87.07 & 90.79 & 52.0 & 56.0 & 64.2 \\ 
    & \CC{20}PaCo + \score-FL (ours) & \CC{20}\textbf{85.61} & \CC{20}\textbf{87.49} & \CC{20}\textbf{91.80} & \CC{20}\textbf{53.71} & \CC{20}\textbf{56.84} & \CC{20}\textbf{65.13} \\ \hline
    One-Stage training & GLMC~\cite{glmc} & 88.50 & 91.04 & 94.90 & 58.0 & 63.78 & 73.43 \\
    & \CC{20}GLMC + \score-GC (ours) & \CC{20}89.38 & \CC{20}90.32 & \CC{20}94.67 & \CC{20}60.01 & \CC{20}63.16 & \CC{20}73.50 \\
    & \CC{20}GLMC + \score-FL (ours) & \CC{20}\textbf{92.33} & \CC{20}\textbf{93.87} & \CC{20}\textbf{94.93} & \CC{20}\textbf{61.33} & \CC{20}\textbf{64.90} & \CC{20}\textbf{73.78} \\
    \bottomrule 
    \end{tabular}}
    \vspace{-2ex}
\end{table*}

\textbf{CIFAR-100-LT} is an extension of~\cite{cifar} with fine-grained labels. CIFAR-100-LT is also created by sampling its balanced counterpart based on an exponentially decaying function (IFs ranging from 10 to 100) but contains several few-shot classes (less than 20 instances per class).\looseness-1

\textbf{ImageNet-LT} introduced in~\cite{oltr}, is a subset of the ImageNet~\cite{imagenet} dataset consisting of 115.8K images from 1000 categories. 
The dataset shows severe imbalance following an exponentially decreasing distribution with a maximum and minimum of 1280 and 5 images per class.\looseness-1

\textbf{MedMNIST}~\cite{medmnist} demonstrates real-world imbalance in medical datasets. We adopt the \textit{OrganAMNIST} and \textit{DermaMNIST} subsets of MedMNIST as they present extreme class-imbalance. 
OrganAMNIST consists of 41072 axial slices from CT volumes, highlighting 11 distinct organ structures while DermaMNIST contains 8010 samples of 7 different varieties of pigmented skin lesions.
The data distributions of the adopted benchmarks are depicted in \cref{fig:data_dist_cifar}(c, d).\looseness-1 

\begin{table}[t]
\caption{\textbf{Longtail Recognition performance (Top-1 Acc \%) on ImageNet-LT dataset}. We show that objectives in \score (shaded in \colorbox{LightGreen}{Green}) generalize to existing SoTA approaches with improved overall performance. \textsuperscript{*} indicates models trained with ResNet-50 as backbone.}
\centering
\small
\resizebox{\columnwidth}{!}{\begin{tabular}{l|ccc|c}
\hline
\multirow{ 2}{*}{\textbf{Method}} & \multicolumn{4}{c}{\textbf{ImageNet-LT}} \\
\cline{2-5}
 &  \textbf{Many} & \textbf{Med} & \textbf{Few} & \textbf{All} \\
\midrule \midrule
CE\textsuperscript{*} \cite{ce_loss_nips}       & 64.0  & 33.8 & 5.8 & 41.6 \\
SupCon\textsuperscript{*} \cite{supcon2020}    & 53.4  & 2.9    & 0.0  & 22.0 \\ \hline
CB-Focal\textsuperscript{*} \cite{cb_focal} & 39.6  & 32.7 & 16.8 & 33.2 \\
LDAM\textsuperscript{*} \cite{ldam}    & 60.4  & 46.9 & 30.7 & 49.8 \\
KCL\textsuperscript{*} \cite{kcl}     & 61.8  & 49.4 & 30.9 & 51.5 \\
TSC\textsuperscript{*} \cite{tsc}      & 63.5  & 49.7 & 30.4 & 52.4 \\
BCL \cite{bcl}                         & 67.9  & 54.2 & \textbf{36.6} & 57.1 \\
RIDE \cite{ride} (3 experts)            & 66.4  & 53.9 & 35.6 & 56.2 \\ \hline
\rowcolor{LightCyan}
PaCo \cite{paco} (400 epochs)      & 63.6  & 55.6 & 34.9 & 55.6 \\
\rowcolor{LightGreen}
PaCo + \score-GC (\textbf{ours})   & \textbf{69.4}  & 44.5 & 16.7 & 50.2  \\
\rowcolor{LightGreen}
PaCo + \score-FL (\textbf{ours})   & 68.9  & \textbf{55.8} & 32.3 & \textbf{57.5} \\ \hline
\rowcolor{LightCyan}
GLMC \cite{glmc}                   & 70.1  & 52.4 & 30.4 & 56.1 \\
\rowcolor{LightGreen}
GLMC + \score-GC (\textbf{ours})   & 68.7  & 52.4 & 31.5 & 55.7  \\
\rowcolor{LightGreen}
GLMC + \score-FL (\textbf{ours})   & \textbf{71.4}  & \textbf{56.2} & \textbf{36.6} & \textbf{59.3} \\  \hline
\end{tabular}}
\vspace{-4ex}
\label{tab:imagenet_lt}
\end{table}
\textbf{IDD}~\cite{idd} depicts $\sim41$K real-world traffic scenarios characterized by longtail imbalance, high traffic density and large variability among object classes~\cite{majee2021fewshot}. IDD consists of $\sim31$K training examples and 10.2K validation images for the object detection task. \cref{fig:idd_overview_imbal} depicts the data distribution of IDD highlighting the imbalance in the dataset.\looseness-1\looseness-1

\textbf{LVIS}~\cite{lvis} dataset encapsulates 1203 commonplace objects from the MS-COCO~\cite{coco} detection dataset (which consisted of 80 total classes) with extreme imbalance among classes. 
Every category in LVIS is assigned a distinct identifier from WordNet~\cite{wordnet}. The training dataset comprises a total of 100,000 images, encompassing 1.3 million instances, and the validation set contains 20,000 images.\looseness-1

For CIFAR-10-LT and CIFAR-100-LT experiments (refer \cref{tab:cifar_lt}) we follow the experimental setup of~\cite{glmc} and adopt a ResNet-32~\cite{resnet} backbone. We adopt the setup of PaCo~\cite{gpaco} for the experiments on ImageNet-LT with a ResNeXt-50~\cite{resnext} backbone. 
Additionally, for contrasting against existing metric/contrastive learners under long-tail settings of CIFAR-10 and MedMNIST (refer \cref{tab:imbal_clf}) we follow the experimental setup of~\cite{supcon2020} with a ResNet-50~\cite{resnet} backbone. 
Finally, we adopt the Faster-RCNN + FPN~\cite{focal_loss} architecture with a ResNet-101 backbone in the Detectron2\footnote{\url{https://github.com/facebookresearch/detectron2}} framework for experiments on IDD and LVIS datasets. 
We train all our models on 2 NVIDIA A6000 GPUs with code released at \url{https://github.com/amajee11us/SCoRe.git}. More details on the datasets, experimental setup and hyper-parameters for individual experiments in section \ref{app:exp_setup} of the appendix.\looseness-1

\begin{table}[t]
      \caption{\textbf{Object detection performance on IDD and LVIS datasets:} Applying our combinatorial objectives on a Faster-RCNN + FPN model produces the best Mean Average Precision ($mAP$) on real-world class-imbalanced settings.\looseness-1}
      \centering
      \resizebox{0.96\columnwidth}{!}{\begin{tabular}{ l | c| c | c | c }
            \hline
            \textbf{Method}                               & \textbf{Backbone and head} & $mAP$  & $mAP_{50}$ & $mAP_{75}$ \\
            \hline \hline
            \multicolumn{5}{c}{\textbf{India Driving Dataset (IDD)}} \\
            \hline
            YOLO -V3\footnotemark \citep{yolo3}           & Darknet-53        & 11.7 & 26.7       & 8.9        \\
            Poly-YOLO\footnotemark[4] \citep{polyyolo}    & SE-Darknet-53     & 15.2 & 30.4       & 13.7       \\
            Mask-RCNN\footnotemark[4] \citep{maskrcnn}    & ResNet-50         & 17.5 & 30.0       & 17.7       \\
            Retina-Net \citep{focal_loss}                  & ResNet-50 + FPN   & 22.1 & 35.7       & 23.0       \\
            Faster-RCNN \citep{faster-rcnn}               & ResNet-101        & 27.7 & 45.4       & 28.2       \\
            Faster-RCNN + FPN                             & ResNet-101 + FPN  & 30.4 & 51.5       & 29.7       \\
            \hline
            Faster-RCNN + SupCon                          & ResNet-101 + FPN  & 31.2 & 53.4       & 30.5       \\
            \rowcolor{LightGreen}
            Faster-RCNN + \score-GC [$C_f$]               & ResNet-101 + FPN  & 33.6 & 56.0       & 34.6       \\
            \rowcolor{LightGreen}
            Faster-RCNN + \score-FL [$S_f/C_f$]   & ResNet-101 + FPN  & \textbf{36.3} & \textbf{59.5} & \textbf{37.1} \\
            \hline
            \multicolumn{5}{c}{\textbf{LVIS Dataset}} \\
            \hline
            Faster-RCNN + FPN                             & ResNet-101 + FPN  & 14.2 & 24.4       & 14.9       \\
            Faster-RCNN + SupCon                          & ResNet-101 + FPN  & 14.4 & 26.3       & 14.3       \\
            \rowcolor{LightGreen}
            Faster-RCNN + \score-GC [$C_f$]               & ResNet-101 + FPN  & 17.7 & 29.1      & 18.3         \\
            \rowcolor{LightGreen}
            Faster-RCNN + \score-FL [$S_f/C_f$]   & ResNet-101 + FPN  & \textbf{19.1} & \textbf{30.5} & \textbf{20.3} \\
            \hline
      \end{tabular}}\\
      \label{tab:idd_od}
      \vspace{-4ex}
\end{table}

\footnotetext{Results are from \cite{polyyolo}.}
\vspace{-2ex}
\subsection{Results on Long-tail Image Classification}
\paragraph{Benchmark Results:} At first, we conduct experiments on the long-tail benchmarks of CIFAR-10-LT, CIFAR-100-LT and ImageNet-LT outlined in~\cite{oltr}.
To the SoTA approaches in long-tail recognition - PaCo and GLMC, we introduce the proposed combinatorial objectives in \score\ as auxiliary supervision head.
For the results on \textbf{CIFAR-10-LT} and \textbf{CIFAR-100-LT} we adopt the benchmark published in GLMC~\cite{glmc} with addition of combinatorial objectives as tabulated in \cref{tab:cifar_lt}. 
We compare objectives in SCoRe against class balancing (row 2,3) / weight balancing techniques (row 4,5), augmentation based approaches (rows 6, 7, 8) and approaches adopting auxiliary supervision (mostly self-supervised) in training (rows 9 - 14). 
We observe that addition of \score\ objectives, specifically \score-FL improves overall performance across various degrees of imbalance (denoted as IF in \cref{tab:cifar_lt}), upto 1.11\% over PaCo and 4.33\% over GLMC (IF=100) for the CIFAR-10-LT dataset. 
In particular, the gain in performance is significant under severe imbalance (IF=100) whereas, in a more balanced setting (IF=10) the performance boost is incremental. 
We show that the aforementioned observation continues to hold in CIFAR-100-LT (containing fine-grained labels and few-shot classes) with 3.29\% gain over PaCo and 5.74\% gain over GLMC (IF=100), where introduction of combinatorial objectives boosts performance in all settings significantly on ones with large imbalance (IF=100).\looseness-1

Secondly, experiments conducted on \textbf{ImageNet-LT} shown in \cref{tab:imagenet_lt} clearly demonstrates the supremacy of objectives in \score\ which demonstrate combinatorial properties by outperforming SoTA approaches like PaCo~\cite{gpaco} and GLMC~\cite{glmc} by 2.1\% (\textit{All} category) and 3.6\% respectively.
Introduction of combinatorial objectives like \score-FL \emph{significantly improves performance on the tail classes without significant loss in performance on head ones} reinstating the inherent \emph{class-balancing} property of \score-FL.
However, we observe a drop in performance in performance in the few-shot (\textit{Few}) and comparable performance in many shot (\textit{Many}) settings. This can be attributed to the objectives guiding the model to overfit on the tail classes during training.
For all datasets in the aforementioned benchmarks, it is interesting to note that \score\ objectives augment the underlying architecture in existing SoTA approaches~\cite{gpaco} thus \emph{demonstrating the generalizability of our objectives} in this domain.\looseness-1

\begin{table}[t]
      \caption{\textbf{Multi-class classification performance (Top1-Accuracy \%) of combinatorial objectives in \score} (shaded in \colorbox{LightGreen}{Green}) against existing approaches in metric learning and their submodular instances (shaded in \colorbox{LightCyan}{blue}) on \textbf{Class-Imbalanced} \textit{CIFAR-10-LT} (columns 2 - 3) and \textit{MedMNIST} (columns 4 - 5) datasets.\looseness-1}
      \centering
      \small
      \setlength{\tabcolsep}{2pt}
       \resizebox{\columnwidth}{!}{\begin{tabular}{l|c|c|c|c}
           \hline
           \multirow{ 3}{*}{\textbf{Objective Function}} & 
           \multicolumn{2}{c|}{\textbf{CIFAR-10-LT}} & 
           \multicolumn{2}{c}{\textbf{MedMNIST}} \\
           \cline{2-5}
           & \textbf{LongTail}& \textbf{Pathological}& \textbf{OrganMNIST} & \textbf{Derma} \\
           & \textbf{IF=10}& \textbf{Step}& \textbf{(Axial)} & \textbf{MNIST}\\
            \hline \hline
            Cross-Entropy (CE)                 & 86.44  & 74.49   & 81.80       & 71.32 \\
            Triplet Loss~\cite{triplet}        & 85.94  & 74.23   & 81.10       & 70.92 \\   
            N-Pairs~\cite{n_pairs}             & 89.70  & 73.10   & 84.84       & 71.82 \\
            Lifted Structure Loss~\cite{lsl}   & 82.86  & 73.98   & 84.55       & 71.62 \\
            SNN~\cite{snn}                     & 83.65  & 75.97   & 83.85       & 71.87 \\
            Multi-Similarity Loss~\cite{msloss}& 82.40  & 76.72   & 85.50       & 71.02 \\
            SupCon~\cite{supcon2020}           & 89.96  & 78.10   & 87.35       & 72.12 \\
            \hline
            \rowcolor{LightCyan}
            Submod-Triplet (ours)              & 89.20  & 74.36   & 86.03       & 72.35 \\
            \rowcolor{LightCyan}
            Submod-SNN (ours)                  & 89.28  & 78.76   & 86.21       & 71.77 \\
            \rowcolor{LightCyan}
            SupCon-Var (ours)               & 90.81  & 81.31   & 87.48       & 72.51 \\
            \hline
            \rowcolor{LightGreen}
            \score-GC [$S_{f}$] (ours)         & 89.20  & 76.89   & 86.28       & 69.10 \\
            \rowcolor{LightGreen}
            \score-GC [$C_{f}$] (ours)         & 90.83  & 87.37   & \textbf{87.57}       & 72.82 \\
            \rowcolor{LightGreen}
            \score-LogDet [$C_{f}$] (ours)            & 90.80  & 87.00   & 87.00       & 72.04 \\
            \rowcolor{LightGreen}
            \score-FL [$C_{f}$/ $S_{f}$] (ours)     & \textbf{91.80}  & \textbf{87.49}   & 87.22       & \textbf{73.77} \\
            \hline
      \end{tabular}}\\
      \vspace{-3ex}
      \label{tab:imbal_clf}
      
\end{table}
\vspace{-2ex}
\paragraph{Generalization to Existing Metric/Contrastive Learners:} \cref{tab:imbal_clf} compares the performance of the newly introduced objectives in \score\ against existing metric/contrastive learners on the framework in~\cite{supcon2020}.
We conduct experiments on the \textit{Longtail} (IF=10) and \textit{step} distributions of the \textbf{CIFAR-10} dataset (refer \cref{app:exp_setup}). 
\begin{figure*}[t]
        \centering
        \includegraphics[width=0.9\textwidth]{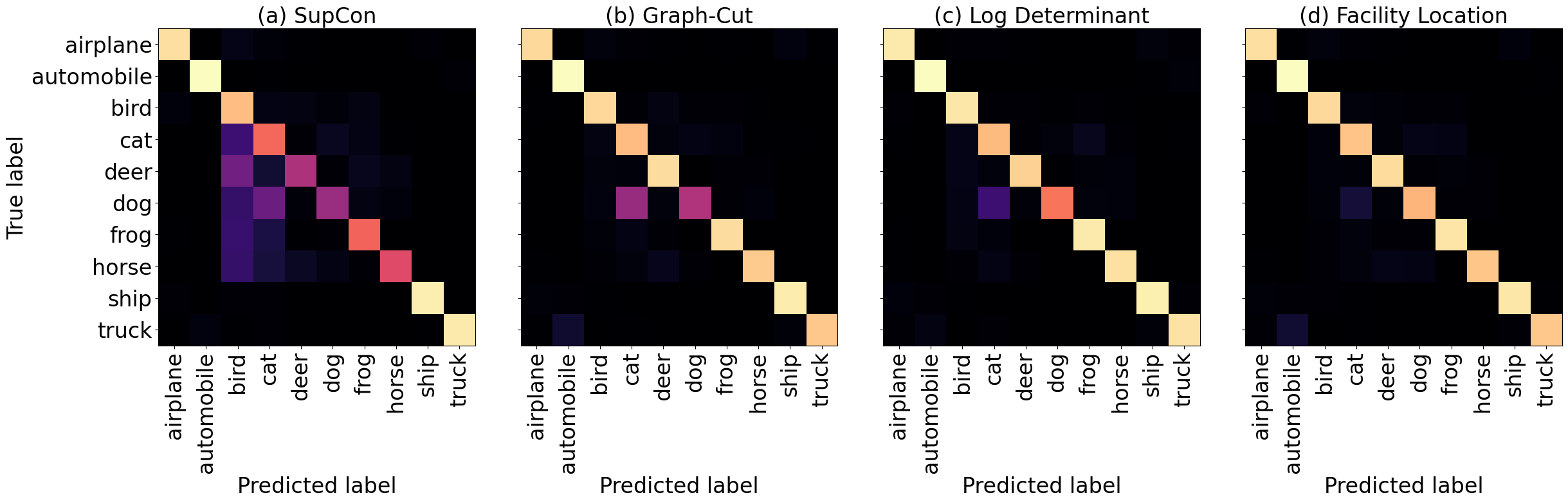}
        \caption{\textbf{Comparison of Confusion Matrix plots} between (a) SupCon \cite{supcon2020}, (b) Graph-Cut (GC), (c) Log Determinant, and (d) Facility Location (FL) for the longtail imbalanced setting of CIFAR-10 dataset. We show a significant reduction in inter-class bias when employing combinatorial objectives in \score\ characterized by reduced confusion between classes.}
        \label{fig:conf_map_cifar}
        \vspace{-2ex}
\end{figure*}
\begin{figure}[ht]
        \centering
        \includegraphics[width=\columnwidth]{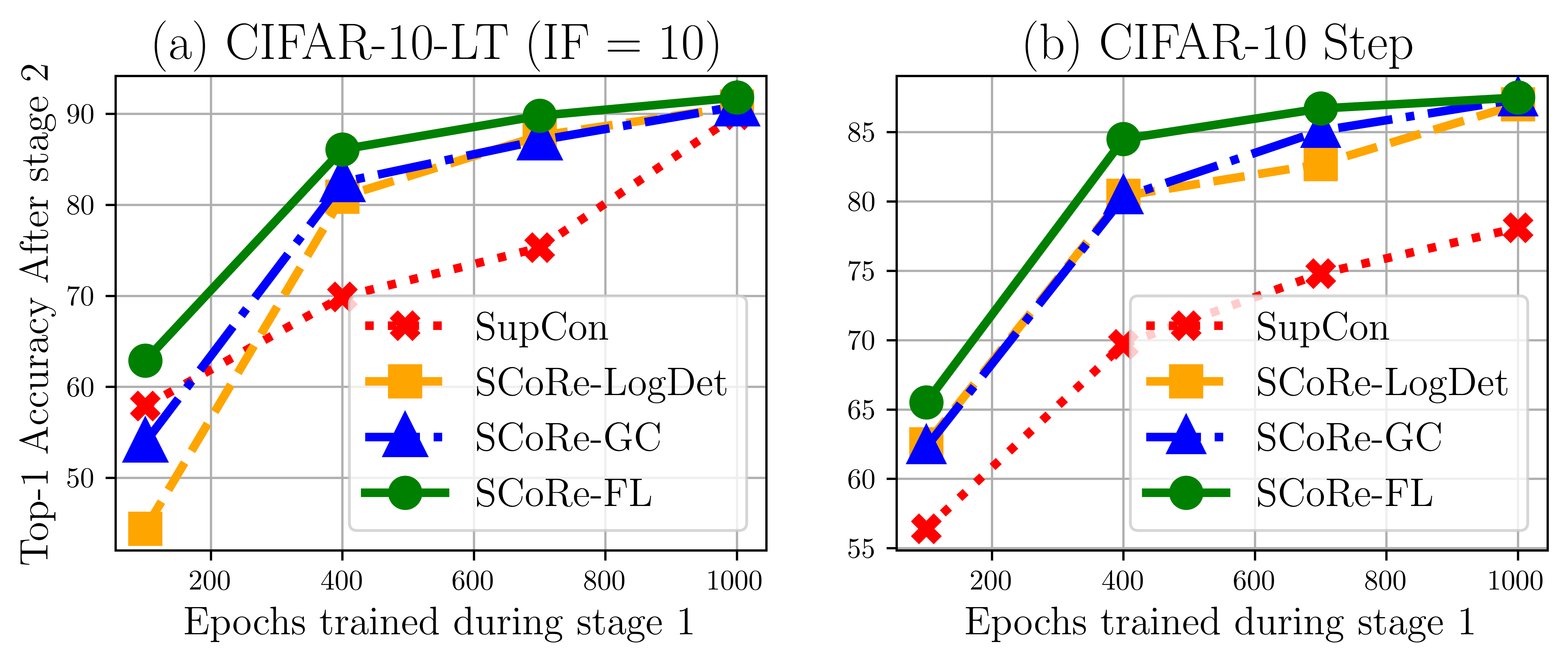}
        \vspace{-2ex}
        \caption{\textbf{\score\ demonstrates faster learning of discriminative representations} in CIFAR-10 with fewer training epochs for both  \textit{longtail} and \textit{step} imbalanced settings.}
        \label{fig:score_convergence}
        \vspace{-4ex}
\end{figure}
The combinatorial objectives in \score, namely the \score-FL objective shows a 2\% and 7.6\% improvement over SupCon for the \textit{longtail} and \textit{step} distributions respectively. 
Even the reformulated submodular objectives - \textit{Submod-Triplet}, \textit{Submod-SNN} and \textit{SupCon-Var} demonstrate upto 3.5\%, 3.7\% and 4.11\% respectively over their non-submodular counterparts.
Further, we demonstrate the effectiveness of the \score\ objectives on the naturally imbalanced \textit{OrganAMNIST} and \textit{DermaMNIST} subsets of \textbf{MedMNIST} in \cref{tab:imbal_clf} (columns 4,5). 
\score\ objectives outperform SoTA approaches by 0.25\% (as shown by \score-GC over SupCon) and 1.5\% (as shown by \score-FL over SupCon) for OrganMNIST and DermaMNIST respectively. 
Similar to CIFAR-10 benchmark we also observe that, submodular counterparts of existing contrastive losses consistently outperform their non-submodular counterparts.\looseness-1
\vspace{-3ex}
\paragraph{Discriminative Clustering and Convergence:} We compare confusion matrix plots on predicted class labels after stage 2 of model training for the CIFAR-10-LT (IF=10) dataset.
Plots in Figure \ref{fig:conf_map_cifar} show that SupCon shows $\sim22$\% overall confusion with elevated confusion between the \textit{animal} hierarchy of CIFAR-10. 
Both \score-GC and \score-LogDet demonstrate confusion between structurally similar objects like \textit{cat} and \textit{dog} (4-legged animals).
Interestingly, a significant drop in confusion is observed when adopting \score-FL with a minimum of 8.2\%.
The reduction in confusion by objectives proposed in \score\ \textbf{points to a reduction in inter-class bias}~\cite{majee2021fewshot}. This is correlated to reducing the impact of class-imbalance due to formation of discriminative feature clusters.\looseness-1

Finally, from \cref{fig:score_convergence} we show that \score\ facilitates the learning \textbf{robust feature representations within significantly fewer training epochs}. Although we see a larger performance gain in the long-tail setting (\score-FL) (\cref{fig:score_convergence}(a,b)), \score\ demonstrates a \emph{faster convergence in the balanced setting} as well as discussed in \cref{app:abl_convergence}.\looseness-1

\subsection{Results on Class-Imbalanced Object Detection}
We benchmark the performance of our approach against SoTA object detectors which adopt Focal Loss~\cite{focal_loss}, data-augmentations etc. 
At first, we introduce a contrastive learning based objective (SupCon) in the box classification head of the object detector and show that contrastive learning outperforms standard model training (using CE loss) on IDD by 3.6 \% (1.9 $mAP_{50}$ points) and 10.6\% (2.8 $mAP_{50}$ points) on LVIS datasets.
Secondly, we introduce the objectives in \score\ to the box classification head and show that they outperform the SoTA as well as the contrastive learning objective~\cite{supcon2020}. The results in Table \ref{tab:idd_od} show that the \score-FL and \score-GC objectives outperform the SoTA method by 6.1 $mAP_{50}$ and 2.6 $mAP_{50}$ points respectively on IDD, alongside 6.1 $mAP_{50}$ and 4.7 $mAP_{50}$ points respectively on LVIS. 
Additionally, from the class-wise performance on IDD as shown in \cref{fig:idd_overview_imbal},  \score\ objectives demonstrate a sharp rise in performance ($mAP_{50}$ value) of the rare classes (a maximum of 6.4 $mAP$ points for \textit{Bicycle} class) over SoTA objectives.\looseness-1
\vspace{-2ex}

\section{Conclusion}
The \score\ framework introduces a novel family of combinatorial objectives based on submodular information measures, designed to address long-tail imbalance in real-world vision tasks. 
A key strength of SCoRe is its ability to present new methodologies alongside generalizing to existing metric and contrastive learners. 
Empirically, SCoRe demonstrates remarkable effectiveness, achieving up to 5.74\% improvement in classification tasks on datasets like CIFAR-10-LT, CIFAR-100-LT and MedMNIST, 2.1\% on ImageNet-LT and an impressive 19.4\% enhancement in object detection on challenging datasets such as the Indian Driving Dataset (IDD) and LVIS. 
The integration of combinatorial counterparts of existing objectives further underscores SCoRe's versatility, leading to significant performance gains and validating its efficacy in managing class imbalance. SCoRe's bridging of novel and established learning strategies marks a substantial contribution to the field, offering a robust solution for real-world applications.\looseness-1

\section*{Impact Statement}
This paper presents work whose goal is to advance the field of Machine Learning in general and Representation Learning in particular. There are many potential societal consequences of our work, none which we feel must be specifically highlighted here.

\section*{Acknowledgements}
We gratefully thank anonymous reviewers for their valuable comments. We would also like to extend our gratitude to our fellow researchers from the CARAML lab at UT Dallas - Nathan Beck and Truong Pham for their suggestions. This work is supported by the National Science Foundation under Grant Numbers IIS-2106937, a gift from Google Research, an Amazon Research Award, and the Adobe Data Science Research award. Any opinions, findings, and conclusions or recommendations expressed in this material are those of the authors and do not necessarily reflect the views of the National Science Foundation, Google or Adobe.

\bibliography{references}
\bibliographystyle{icml2024}

\newpage
\appendix
\onecolumn
\section{Appendix}
\subsection{Notations}
\label{app:notation}
Following the problem definition in Section \ref{sec:method} we introduce several important notations in Table \ref{tab:notations} that are used throughout the paper. 
\begin{table*}[ht]
      \caption{Collection of notations used in the paper.}
      \centering
      \begin{tabular}{ c | c }
            \hline
           \textbf{Symbol}  & \textbf{Description} \\
            \hline
            $\mathcal{T}$ & The training Set. $|\mathcal{T}|$ denotes the size of the training set. \\
            $\mathcal{V}$ & Ground set containing feature vectors from all classes in $\mathcal{T}$. \\
            $F(x, \theta)$ & Convolutional Neural Network used as feature extractor. \\
            $Clf(.,.)$ & Multi-Layer Perceptron as classifier. \\
            $\theta$ & Parameters of the feature extractor. \\
            $S_{ij}(\theta)$ & Similarity between images $i,j \in \mathcal{T}$. \\
            $D_{ij}(\theta)$ & Distance between images $i,j \in \mathcal{T}$. \\
            $p$ & Positive sample which is of the same class $c_i$ as the anchor $a$. \\
            $n$ & Negative sample which is of the same class $c_i$ as the anchor $x$. \\   
            $A_k$ & Target set containing feature representation from a single class $k \in c_i$.\\
            $f(A)$ & Submodular Information function over a set $A$. \\
            $S_{f}$ & Variant of submodular information function denoting total information in the ground set $V$. \\
            $C_{f}$ & Variant of submodular information function denoting total correlation in the ground set $V$. \\
            $L(\theta)$ & Loss value computed over all classes $c_i \in C$. \\
            $f(A_k,\theta)$ & Instantiation of objective functions in \score\ over a set/class $A_k$ given parameters $\theta$. \\ \hline
            AK & Actinic Keratoses \\
            BCC & Basal Cell Carcinoma \\
            KLL & Keratosis-Like-Lesions \\
            DF & Dermatofibroma \\
            M & Melanoma \\
            MN & Melanocytic Nevi \\
            VL & Vascular Lesions \\            
            \hline
      \end{tabular}\\
      \label{tab:notations}
      
\end{table*}

\subsection{Experimental Setup : Additional Information}
In this section we iron out the dataset details, training and inference settings of various datasets/tasks encompassed in the \score\ framework.
\label{app:exp_setup}
\subsubsection{Class-Imbalanced Image Classification on Cannonical benchmarks}
\begin{figure*}[t]
        \centering
        \includegraphics[width=0.9\textwidth]{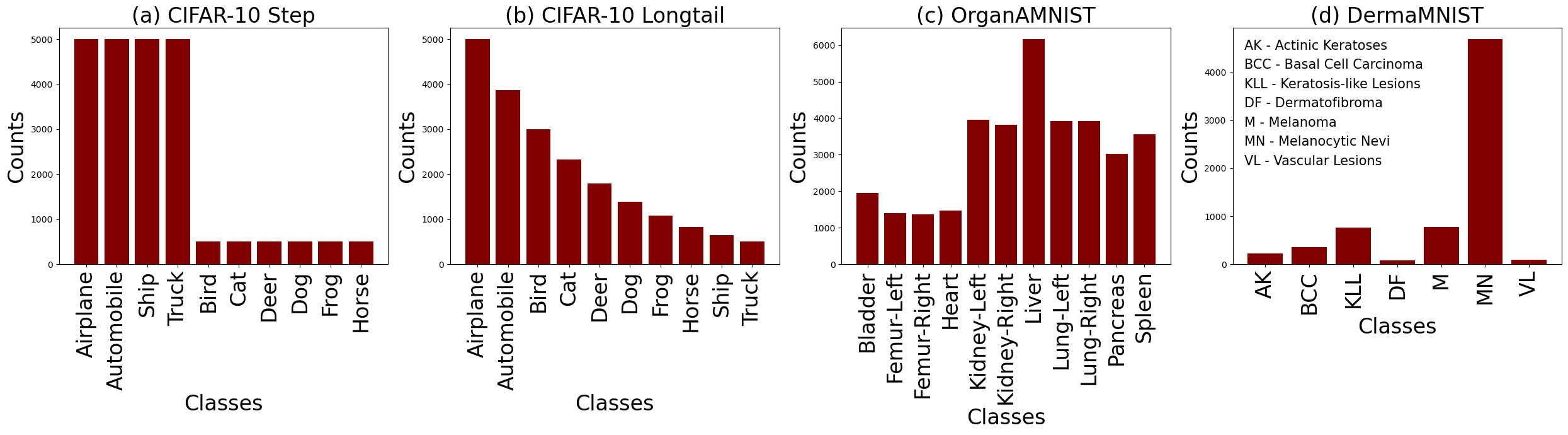}
        \vspace{-3ex}
        \caption{\textbf{Data Distribution of CIFAR-10 and MedMNIST datasets} under the class-imbalanced setting. Subfigure (a,b) depicts the longtail and step based pathological class imbalanced settings in CIFAR-10 and (c,d) depicts the naturally imbalanced OrganAMNIST and DermaMNIST\footnotemark datasets respectively.}
        \label{fig:data_dist_cifar}
        \vspace{-2ex}
\end{figure*}

\paragraph{CIFAR-10-LT and CIFAR-100-LT:} Following the 
discussion on the choices of datasets introduced by the OLTR~\cite{oltr} benchmark in \cref{sec:experiments} we vary the Imbalance Factors (IF) of an exponentially decaying function to sample the CIFAR-10 and CIFAR-100 datasets to create their respective long-tail counterparts.
We vary the imbalance factors between 10, 50 and 100 to produce pathologically imbalanced datasets. The higher the value of IF, the larger is the number of tail classes a sample of which for IF=10 is depicted in \cref{fig:data_dist_cifar}.
Additionally we introduce a \textbf{step} function based imbalance setting which exploits the hierarchy already available in CIFAR-10. The CIFAR-10 dataset can be broadly classified into \textit{animal} and \textit{automobile} classes.
We use this information to subsample the \textit{animal} (chosen at random) class objects to create an imbalanced step data distribution. The distributions of the dataset is depicted in Figure \ref{fig:data_dist_cifar}. 

For contrasting against SoTA metric/contrastive learners in \cref{tab:imbal_clf}, we train our models by adopting the training strategy of \cite{supcon2020} and release the codebase at \url{https://github.com/amajee11us/SCoRe.git}. For stage 1 we train a ResNet-50 backbone with a batch size of 512 (1024 after augmentations) with an initial learning rate of 0.4, trained for 1000 epochs with a cosine annealing scheduler and a temperature for the combinatorial objectives to be 0.7.
In stage 2 we freeze the backbone and use the output of the final pooling layer to train a linear classifier $Clf$ with a batch size of 512 and a constant learning rate of 0.8. 

Further, for contrasting against existing SoTA approaches in Long-Tail recognition as shown in \cref{tab:cifar_lt} for both CIFAR-10 and CIFAR-100 datasets, we introduce the objectives defined in \score\ into the existing frameworks of SoTA approaches like GLMC~\cite{glmc} and PaCo~\cite{paco}. For PaCo we replace its contrastive objective - MoCo~\cite{moco} with the combinatorial objectives in \score, while for GLMC we introduce our novel objectives to contrast between local and global features in the original architecture.

\paragraph{ImageNet-LT}: Unlike CIFAR-10, ImageNet~\cite{imagenet} dataset presents a large scale image recognition benchmark. Similar to CIFAR-10 authors in \cite{oltr} subsample this dataset to introduce a pathological imbalance. In contrast to CIFAR-10 ImageNet contains 1000 classes and the longtail version ImageNet-LT contains severe imbalance with a maximum of 1280 and minimum of 5 instances for a particular class in the dataset\footnote{The dataset as been adapted from \url{https://liuziwei7.github.io/projects/LongTail.html}.}.

For the ImageNet-LT dataset we follow the training and inference strategy adopted by PaCo~\cite{gpaco} with modifications to the objective functions which are released at \url{https://github.com/amajee11us/SCoRe/blob/main/objectives/combinatorial/PaCoFL.py} (for \score-FL loss function). Unlike the CIFAR-10 dataset PaCo adopts a one stage training strategy where we train the model for 400 epochs on a ResNeXt-50~\cite{resnext} backbone with an initial learning rate of 0.1 and a cosine annealing scheduler. The model is trained on 4 GPUs, with a batch size of 32 on each GPU and a temperature for the combinatorial objectives to be 0.7.
The results on the longtail distribution of ImageNet-LT benchmarked against several approaches in Longtail learning have been depicted in \cref{tab:imagenet_lt}.

\subsubsection{Class-Imbalanced Medical Image Classification} 
In contrast to pathological imbalance introduced in cannonical benchmarks we conduct our experiments on two subsets of \textbf{MedMNIST} \cite{medmnist} dataset which demonstrate a natural class-imbalanced setting. \\

\textbf{OrganAMNIST} dataset consists of axial slices from CT volumes, highlighting 11 distinct organ structures for a multi-class classification task. Each image is of size $[1 \times 28 \times 28]$ pixels. The OrganAMNIST dataset contains 34581 training and 6491 validation samples of single channel images highlighting various modalities of 8 different organs.\\

\textbf{DermaMNIST} subset presents dermatoscopic images of pigmented skin lesions, also resized to $[3 \times 28 \times 28]$ pixels. This dataset supports a multi-class classification task with 7 different dermatological conditions. Although the DermaMNIST has RGB images, it is a small scale dataset with a total of 7007 training samples and 1003 validation samples.

Similar to CIFAR-10 we train our models on MedMNIST datasets using the two stage training strategy outlined in \score\ at \url{https://github.com/amajee11us/SCoRe.git}.
For both these subsets used in our framework, pixel values were normalized to the range \([0,1]\), and we relied on the standard train-test splits provided with the datasets for our evaluations. The results from the experiments are discussed in Section \ref{res:classification}. For both data subsets we train the model for 500 epochs in stage-1 with a ResNet-50 backbone, a batch size of 256 and a cosine annealing scheduler. For stage-2 we use the frozen feature extractor and train a classifier with a batch size of 128 for 100 epochs with early stopping.

\footnotetext{Abbreviations are included in section \ref{app:notation} of the appendix.}

\subsubsection{Class-Imbalanced Object Detection}  
\begin{figure*}[t]
        \centering
        \includegraphics[width=0.95\textwidth]{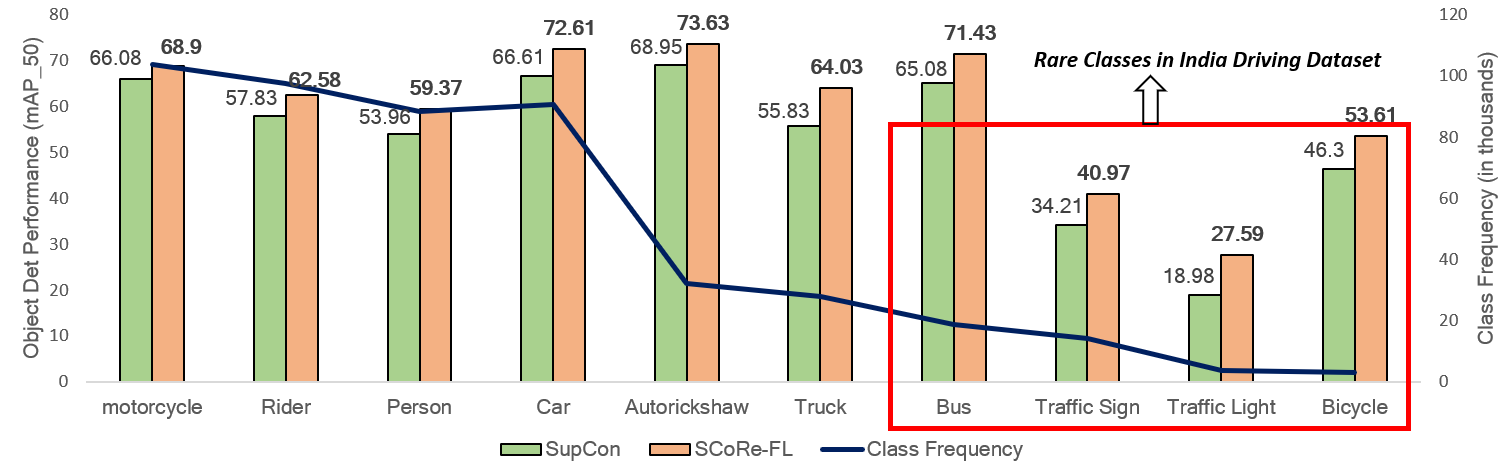} 
        \caption{\textbf{The effect of class-imbalance} on the performance metrics ($mAP_{50}$) for the object detection task of the India Driving Dataset (IDD). As the class frequency (shown as \textcolor{blue}{blue} line) decreases, proposed objectives in \score\ (shown in \textcolor{red}{red}) consistently outperforms SoTA approaches like SupCon (shown in \textcolor{green}{green}) in detecting rare road objects like \textit{bicycle}, \textit{traffic light} etc. in IDD.\looseness-1}
        \label{fig:idd_overview_imbal}
        \vspace{-4ex}
\end{figure*}
\paragraph{IDD-Detection}~\cite{idd} dataset demonstrates an unconstrained driving environment, characterized by natural class-imbalance, high traffic density and large variability among object classes. This results in the presence of rare classes like \textit{autorickshaw}, \textit{bicycle} etc. and small sized objects like \textit{traffic light}, \textit{traffic sign} etc. There are a total of ~31k training images in IDD and ~10k validation images of size $[3 \times 1920 \times 1080]$ with high traffic density, occlusions and varying road conditions. 

The architecture of the object detector is a Faster-RCNN \cite{faster-rcnn} model with a ResNet-101 backbone alongside the Feature Pyramidal Network (FPN) as in \citep{focal_loss} to handle varying object sizes. 
Our framework also draws inspiration from FSCE~\cite{fsce} with proposed modifications to the objective functions. We initialize our Faster-RCNN + FPN model with pretrained weights from a ImageNet trained model and fine-tune the model on IDD / LVIS (full dataset). 
We keep the Region Proposal Network (RPN) and the ROI pooling layers unfrozen to adapt to the rare classes. We also double the maximum number of proposals kept after Non-Maximal Suppression (NMS), bringing more proposals from rare classes to the foreground.
We consider only half the number of proposals from the ROI pooling layer (top 256 out of 512) for computing the loss function. This forces the objective function to better penalize the object detector for predicting low objectness scores for objects belonging to the rare classes.
The model is trained for 17000 iterations with a batch size of 8 and an initial learning rate of 0.02. A step based learning rate scheduler is adopted to reduce the learning rate by 10x at 12000 and 15000 iterations respectively. The results for the class-wise performance on the IDD dataset is depicted in \cref{fig:idd_overview_imbal}.

\paragraph{LVIS}~\cite{lvis} dataset depicts an extreme case of longtail imbalance with a large number of tail classes. The dataset consists of 1203 classes created by extending the label set in MS-COCO~\cite{coco} (consisting of just 80 classes). We adopt the version v1.0 of LVIS for our experiments and conduct our experiments on Faster-RCNN+FPN architecture adopting a ResNet-101 backbone with a batch size of 16, initial learning rate of 0.06 and repeat factor sampling for a total of 180,000 iterations. Similar to IDD , the step based learning rate scheduler is adopted to reduce the learning rate by 1/10 at regular intervals. 

\begin{figure*}
\centering
\subfigure[Inter-Class Bias]{
    \includegraphics[width=\linewidth]{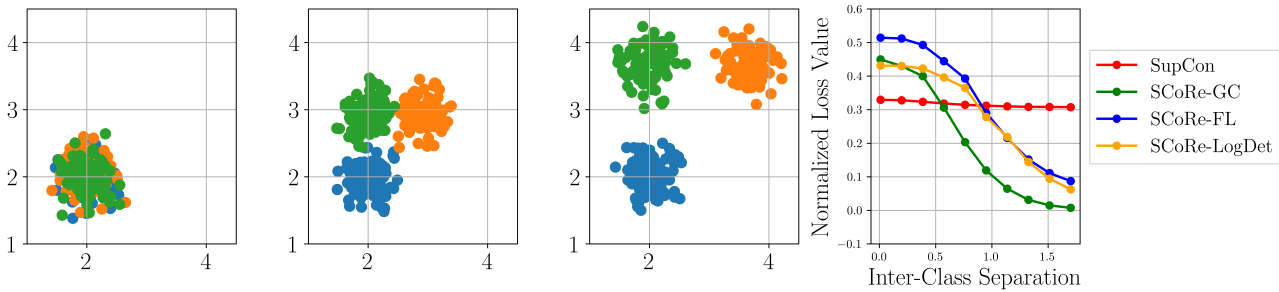}
    \label{fig:inter_class_bias}
} \\
\subfigure[Intra-Class Variance]{
    \includegraphics[width=\linewidth]{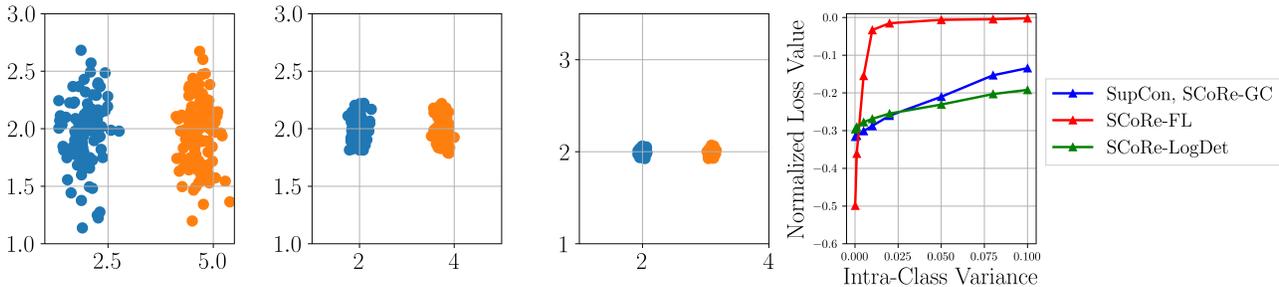}
    \label{fig:intra_class_var}
}
\vspace{-4ex}
\caption{\textbf{Resilience of instantiations in \score\ towards Intra-Class Variance (a) and Inter-Class Bias (b) in representation learning tasks demonstrated through pathologically created synthetic benchmarks.}}
\label{fig:synthetic_loss}
\vspace{-2ex}
\end{figure*}
\subsection{Experiments on Synthetic Datasets}
\label{res:synthetic}
The aim of the experiments on synthetic datasets in \score\ is to demonstrate the variation of \score\ objectives to - \emph{inter-class bias and intra-class variance} and \emph{class-imbalance in long-tail settings}.

To demonstrate resilience to inter-class bias we vary the inter-cluster separation between three disjoint clusters with constant variance of 0.05 as shown in \cref{fig:synthetic_loss}. In \cref{fig:synthetic_loss}(a) we plot the variation of the losses introduced in \cref{tab:comb_overview} to inter-class bias arising to increased overlaps between feature clusters. \textbf{We observe a large variation to inter-class bias in \score\ objectives over existing SoTA methods} demonstrating their resilience to the phenomenon (since the loss would be minimized during back-propagation).

To demonstrate resilience to intra-class variance, we keep the inter-cluster distance (separation between the extremities of two clusters) between two clusters to be constant and vary the variance within each cluster in the range of [0.01, 0.1]. Similar to above we plot the variation of losses in \cref{tab:comb_overview} through \cref{fig:synthetic_loss}(b). The plots show that \score\ objectives, especially \score-FL \textbf{demonstrates a relatively large variation to increase in intra-class variance over existing SoTA approaches} like SupCon~\cite{supcon2020} establishing combinatorial objectives to be resilient to the phenomenon.

Note, that for our experiments all synthetic data-points lie in the first quadrant and thus we adopt the Radial Bias Field (RBF) kernel \cite{pmlr-v139-killamsetty21a} as the similarity metric for computing $S_{ij}$.

\begin{wrapfigure}{r}{0.45\textwidth}
        \centering
        \includegraphics[width=0.45\textwidth]{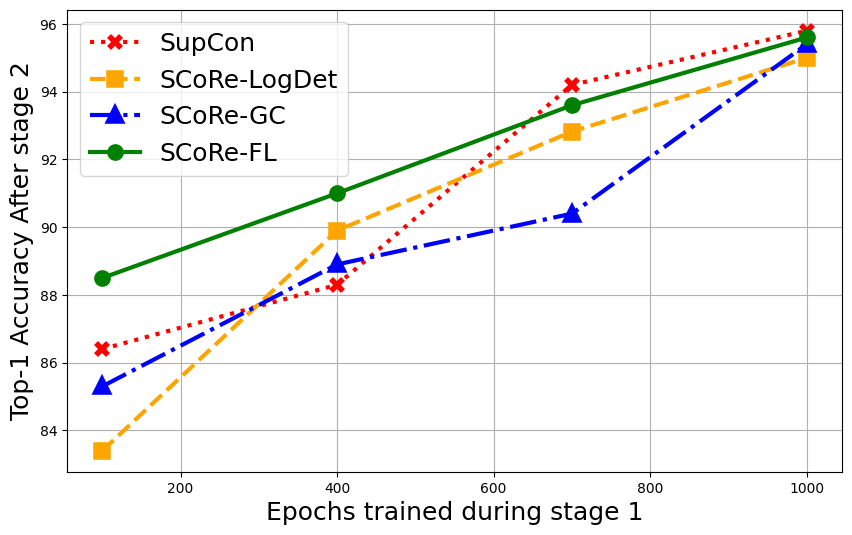}
        \vspace{-2ex}
        \caption{\textbf{Ablation Study}: \score\ demonstrates faster learning of discriminative representations even on the \textit{balanced} setting of CIFAR-10 dataset.}
        \label{fig:abl_balanced_convergence}
        \vspace{-4ex}
\end{wrapfigure}

\subsection{Ablation Study : \score\ Objectives Learn Discriminative Features with Fewer Training Rounds}
\label{app:abl_convergence}
In this experiment we compare the number of training epochs in stage 1 required to learn discriminative feature representations for various instances of \score\ against SoTA contrastive learner (SupCon). 
In stage 1 we train our models (with varying objectives) for various number of epochs in the range of [100, 1000] on the balanced and imbalanced settings of the CIFAR-10 dataset. 
We keep the remaining hyperparameters constant by training our models with a initial learning rate of 0.4, batch size of 512, temperature 0.7 and a cosine annealing scheduler. 
For each of these trained models from stage 1 we train the classifier $Clf$ in stage 2 until convergence and report the Top-1 Accuracy \%. 
Although, the performance (Top-1 Accuracy \%) of our models in balanced setting is lower than the SoTA contrastive learners we observe in \cref{fig:abl_balanced_convergence} that combinatorial objectives like \score-FL learn robust representations within very few training rounds in stage 1 of model training as compared to SoTA approach SupCon~\cite{supcon2020}.
Interestingly, in the imbalanced setting as shown in \cref{fig:score_convergence} models trained with combinatorial objectives in \score\ (\score-FL, \score-GC and \score-LogDet) learn better discriminative features within fewer number of epochs in stage 1. 
This allows model developers to rapidly train generalizable feature extractors for several downstream applications. 

\begin{wraptable}{r}{5.5cm}
    \caption{\textbf{Ablation study} on the effect of $\lambda$ on the performance of Graph-Cut based combinatorial objective in \score.}
    \centering
        \begin{tabular}{c|c}
        \hline
        \multirow{ 2}{*}{$\lambda$} & Top-1 acc\\
                                    & CIFAR-10 (longtail)\\ \hline
        0.5                & 83.65               \\
        \textbf{1.0}       & \textbf{89.96}      \\
        1.5                & 87.11               \\
        2.0                & 85.86               \\ \hline
        \end{tabular}
    \label{tab:abl_lambda}
    \vspace{-4ex}
\end{wraptable}

\subsection{Ablation Study : Effect of $\lambda$ on performance of Graph-Cut based Objective}
In this section we perform experiments on the hyperparameter $\lambda$ introduced in Graph-Cut based combinatorial objective in \score.
The hyper-parameter $\lambda$ is applied to the sum over the penalty associated with the positive set forming tighter clusters. This parameter controls the degree of compactness of the feature cluster ensuring sufficient diversity is maintained in the feature space. For \score-GC to be submodular it is also important for $\lambda $to be greater than or equal to 1 ($\lambda \geq 1$). 
For this experiment we train the two stage framework in \score\ on the longtail CIFAR-10 dataset for 500 epochs in stage 1 with varying $\lambda$ values in range of [0.5, 2.0] and report the top-1 accuracy after stage 2 model training on the validation set of CIFAR-10.
Table \ref{tab:abl_lambda} shows that we achieve highest performance for $\lambda = 1$ for longtail image classification task on the CIFAR-10 dataset. We adopt this value for all experiments conducted on GC in this paper.

\subsection{Proof of theorems of Submodular Combinatorial Objectives}
\label{app:proof_submod_loss}
\subsubsection{Facility Location}
In this section we show proofs for the introduced $L_{S_f}$ and $L_{C_f}$ for the facility location function in \cref{thm:fl}.
\vspace{-2ex}
\begin{proof}
The facility location function over a set $A$ can be represented as $f(A) = {\underset{i \in \mathcal{V}}{\sum}} {\underset{j \in A}{\max}} S_{ij}$. \cref{thm:fl} instantiates this function in \score\ to define two combinatorial objectives $L_{S_f}$ and $L_{C_f}$ as presented by \cref{eq:fl} in the main paper.

From the definition of $L_{S_f}$, the \score-FL ($L_{S_f}$) objective can be derived as $L_{S_f}(\theta) = \overset{|C|}{\underset{k=1}{\sum}} \frac{1}{|\mathcal{V}|} f(A_k, \theta)$. Substituting the instance of FL $f(A_k, \theta) = {\underset{i \in \mathcal{V}}{\sum}} {\underset{j \in A_k}{\max}} S_{ij}$ in the equation we get:
\begin{align*}
        L_{S_f}(\theta) &= \frac{1}{|\mathcal{V}|}\overset{|C|}{\underset{k=1}{\sum}} f(A_k, \theta) \\
                        &= \frac{1}{|\mathcal{V}|}\overset{|C|}{\underset{k=1}{\sum}} {\underset{i \in \mathcal{V}}{\sum}} {\underset{j \in A_k}{\max}} S_{ij} \\
                        &= \frac{1}{|\mathcal{V}|}\overset{|C|}{\underset{k=1}{\sum}} {\underset{i \in \mathcal{V} \setminus A_k}{\sum}} {\underset{j \in A_k}{\max}} S_{ij} + \overset{|C|}{\underset{k=1}{\sum}} {\underset{i \in A_k}{\sum}} {\underset{j \in A_k}{\max}} S_{ij} \\
        L_{S_f}(\theta) &= \frac{1}{|\mathcal{V}|}\overset{|C|}{\underset{k=1}{\sum}} {\underset{i \in \mathcal{V} \setminus A_k}{\sum}} {\underset{j \in A_k}{\max}} S_{ij} + |\mathcal{V}| , \, \, \text{since ${\underset{i \in A_k}{\sum}} {\underset{j \in A_k}{\max}} S_{ij}$ is a constant over the set $A_k$} \\                        
\end{align*}
Now, considering the $C_f$ formulation in \cref{eq:score_formulation} and substituting $f(A_k, \theta)$ we get:
\begin{align*}
    L_{C_f}(\theta) =& \frac{1}{|\mathcal{V}|}\Biggl[\overset{|C|}{\underset{k=1}{\sum}} f(A_k, \theta) - f(\overset{|C|}{\underset{k=1}{\cup}}A_k, \theta)\Biggr]\\
    L_{C_f}(\theta) =& L_{S_f}(\theta) - \frac{1}{|\mathcal{V}|}f(\overset{|C|}{\underset{k=1}{\cup}}A_k, \theta)
\end{align*}
We know that $\overset{|C|}{\underset{k=1}{\cup}} A_k = \mathcal{V}$. Thus,
\begin{align*}
            L_{C_f}(\theta) =& \frac{1}{|\mathcal{V}|}\overset{|C|}{\underset{k=1}{\sum}} {\underset{i \in \mathcal{V} \setminus A_k}{\sum}} {\underset{j \in A_k}{\max}} S_{ij} + |\mathcal{V}| - f(\mathcal{V}, \theta)\\
            =& \frac{1}{|\mathcal{V}|}\overset{|C|}{\underset{k=1}{\sum}} {\underset{i \in \mathcal{V} \setminus A_k}{\sum}} {\underset{j \in A_k}{\max}} S_{ij} + |\mathcal{V}| - {\underset{i \in \mathcal{V}}{\sum}} {\underset{j \in \mathcal{V}}{\max}} S_{ij} \\
            =& \frac{1}{|\mathcal{V}|}\overset{|C|}{\underset{k=1}{\sum}} {\underset{i \in \mathcal{V} \setminus A_k}{\sum}} {\underset{j \in A_k}{\max}} S_{ij}, \, \, \text{since ${\underset{i \in \mathcal{V}}{\sum}} {\underset{j \in \mathcal{V}}{\max}} S_{ij}$ is a constant over the ground set $\mathcal{V}$}
    \end{align*}
Thus, we show that both $L_{S_f}$ as well as $L_{C_f}$ versions of the Facility-Location (\score-FL) function introduced in \cref{thm:fl} are instances of the \score\ framework.    
\end{proof}

\subsubsection{Graph-Cut}
In this section we show proofs for the introduced $L_{S_f}$ and $L_{C_f}$ for the Graph-Cut function in \cref{thm:gc}.
\begin{proof}
The Graph-Cut function over a set $A$ can be represented as $f(A) = {\underset{i \in A, j\in V \setminus A}{\sum}} S_{ij}(\theta) - \lambda \sum_{i, j \in A} S_{ij}(\theta)$. \cref{thm:gc} instantiates this function in \score\ to define two combinatorial objectives $L_{S_f}$ and $L_{C_f}$ as presented by \cref{eq:gc} in the main paper.

From the definition of $L_{S_f}$, the \score-GC ($L_{S_f}$) objective can be derived as $L_{S_f}(\theta) = \overset{|C|}{\underset{k=1}{\sum}} \frac{1}{|A_k|}f(A_k, \theta)$. Substituting the instance of GC $f(A_k, \theta)$ in the equation we get:    
\begin{align*}
        L_{S_f}(\theta) &= \overset{|C|}{\underset{k=1}{\sum}} \frac{1}{|A_k|} f(A_k, \theta) \\
                        &= \overset{|C|}{\underset{k=1}{\sum}} \frac{1}{|A_k|} {\underset{i \in A_k, j\in V}{\sum}} S_{ij}(\theta) - \frac{\lambda}{|A_k|} \sum_{i, j \in A_k} S_{ij}(\theta) \\ 
                        &= \overset{|C|}{\underset{k=1}{\sum}} \frac{1}{|A_k|} {\underset{i \in A_k, j\in V \setminus A_k}{\sum}} S_{ij}(\theta) + \frac{1}{|A_k|}\overset{|C|}{\underset{k=1}{\sum}} {\underset{i \in A_k, j\in A_k}{\sum}} S_{ij}(\theta) - \frac{\lambda}{|A_k|} \sum_{i, j \in A_k} S_{ij}(\theta) \\
\end{align*}
Here, the term $\overset{|C|}{\underset{k=1}{\sum}} {\underset{i \in A_k, j\in A_k}{\sum}} S_{ij}(\theta)$ represents a sum of pairwise similarities over all sets in $\mathcal{V}$. Thus, its value is a constant for a fixed training/ evaluation dataset. Using this condition and ignoring the constant term, we can show that :
\begin{align*}
        L_{S_f}(\theta) &= \overset{|C|}{\underset{k=1}{\sum}} \frac{1}{|A_k|} \Biggl[{\underset{i \in A_k, j\in V \setminus A_k}{\sum}} S_{ij}(\theta) - \lambda \sum_{i, j \in A_k} S_{ij}(\theta)\Biggr] \\
\end{align*}
Now, considering the $C_f$ formulation in \cref{eq:score_formulation} and substituting $f(A_k, \theta)$ we get:
\begin{align*}
    L_{C_f}(\theta) =& \overset{|C|}{\underset{k=1}{\sum}} \frac{1}{|A_k|} f(A_k, \theta) - f(\overset{|C|}{\underset{k=1}{\cup}}A_k, \theta)\\
    L_{C_f}(\theta) =& L_{S_f}(\theta) - \frac{1}{|A_k|}f(\overset{|C|}{\underset{k=1}{\cup}}A_k, \theta) \\
\text{We know that $\overset{|C|}{\underset{k=1}{\cup}} A_k = \mathcal{V}$. Thus,} \\
L_{C_f}(\theta) =& \sum_{k=1}^{|C|} \frac{1}{|A_k|} \left( \sum_{i \in A_k, j \in V \setminus A_k} S_{ij}(\theta) - \lambda \sum_{i, j \in A_k} S_{ij}(\theta) \right) - \\
&\frac{1}{|A_k|} \left( \sum_{i \in \bigcup_{k=1}^{|C|} A_k, j \in V \setminus \bigcup_{k=1}^{|C|} A_k} S_{ij}(\theta) - \lambda \sum_{i, j \in \bigcup_{k=1}^{|C|} A_k} S_{ij}(\theta) \right)
\end{align*}
Since the sets $A_k$ are disjoint, we can simplify the expression as :
\begin{align*}
\sum_{k=1}^{|C|} \frac{1}{|A_k|}\Biggl[\sum_{i, j \in A_k} S_{ij}(\theta) = \sum_{i, j \in \bigcup_{k=1}^{|C|} A_k} S_{ij}(\theta)\Biggr]   
\end{align*}
This leads to a cancellation of the terms involving $\lambda$ in $L_{C_f}(\theta)$:
\begin{align*}
    L_{C_f}(\theta) = \sum_{k=1}^{|C|} \frac{1}{|A_k|} \sum_{i \in A_k, j \in V \setminus A_k} S_{ij}(\theta)
\end{align*}
\end{proof}
Thus, we show that both $L_{S_f}$ as well as $L_{C_f}$ versions of the Graph-Cut (\score-GC) function introduced in \cref{thm:gc} are instances of the \score\ framework. In the final equation shown in \cref{sec:score} we multiply $L_{C_f}$ with a hyper-parameter $\lambda$ to control the penalization of the cross similarity between sets $A_k$ and $\mathcal{V} \setminus A_k$.

\subsubsection{Log-Determinant}
In this section we show proofs for the introduced $L_{S_f}$ and $L_{C_f}$ for the LogDet function in \cref{thm:logdet}.
\begin{proof}
    The submodular function $f(A_k, \theta)$ for Log-Determinant is denoted as $\log\det(S_{A_k} + \lambda\mathbb{I}_{|\mathcal{A_k}|})$ with $\lambda\mathbb{I}$ as the identity term used for numerical stability (empirically).
    Substituting $f(A_k, \theta)$ in the $L_{S_f}$ formulation in \cref{eq:score_formulation} we get:
    \begin{align*}
        L_{S_f}(\theta) &= \overset{|C|}{\underset{k=1}{\sum}} \frac{1}{|A_k|} f(A_k, \theta) \\
                        &= \overset{|C|}{\underset{k=1}{\sum}} \frac{1}{|A_k|} \log \det (S_{A_k}(\theta) + \lambda \mathbb{I}_{|A_k|})  
    \end{align*}
    This leads to the $S_f$ formulation of the LogDet objective in \score.
    Now, considering the $C_f$ formulation in \cref{eq:score_formulation} and substituting $f(A_k, \theta)$ we get:
    \begin{align*}
        L_{C_f}(\theta) =& \overset{|C|}{\underset{k=1}{\sum}} \frac{1}{|A_k|} f(A_k, \theta) - f(\overset{|C|}{\underset{k=1}{\cup}}A_k, \theta)\\
        L_{C_f}(\theta) =& L_{S_f}(\theta) - \frac{1}{|A_k|}f(\overset{|C|}{\underset{k=1}{\cup}}A_k, \theta)
    \end{align*}
    We know that $\overset{|C|}{\underset{k=1}{\cup}} A_k = \mathcal{V}$. Thus,
    \begin{align*}
            L_{C_f}(\theta) =& \overset{|C|}{\underset{k=1}{\sum}} \frac{1}{|A_k|}\log \det (S_{A_k}(\theta) + \lambda \mathbb{I}_{|A_k|}) - \frac{1}{|A_k|}f(\mathcal{V}, \theta)\\
            =& L_{S_f}(\theta) - \frac{1}{|A_k|}\log \det (S_{\mathcal{V}}(\theta) + \lambda \mathbb{I}_{|\mathcal{V}|}) \\
            =& \overset{|C|}{\underset{k=1}{\sum}} \frac{1}{|A_k|}\Biggl[\log \det (S_{A_k}(\theta) + \lambda \mathbb{I}_{|A_k|}) - \log \det (S_{\mathcal{V}}(\theta) + \lambda \mathbb{I}_{|\mathcal{V}|})\Biggr]
    \end{align*}
Thus, we show that both $L{S_f}$ as well as $L_{C_f}$ versions of the Log-Determinant (LogDet) function introduced in \cref{thm:logdet} are instances of the \score\ framework.
\end{proof}

\subsection{Proof of Submodularity for Existing Metric/Contrastive Learners}
\label{app:proof_submod}
In this section we discuss in depth the submodular counterparts of three existing objective functions in contrastive learning. 
We provide proofs that these functions are non-submodular in their existing forms and can be reformulated as submodular objectives through modifications without changing the characteristics of the loss function.

\subsubsection{Triplet loss and Submod-Triplet loss}
\label{app:proof_submod_triplet}
\begin{theorem}
\label{thm:triplet}
    The Triplet loss $L(\theta)$, depicted in row 2 of Table \ref{tab:comb_overview} is not an instance of \score\ in its original form while the slightly modified form, Submod-Triplet $L(\theta) = \sum_{k=1}^{|C|} \frac{1}{|A_k|} \sum_{\substack{i \in A_{k} \\ n \in \mathcal{V} \setminus A_{k}}} S_{in}^{2}(\theta) - \sum_{i,p \in A} S_{ip}^{2}(\theta)$ is an instance of \score\ ($S_f$ version) with the submodular function $f(A, \theta) = \sum_{\substack{i \in A \\ n \in \mathcal{V} \setminus A}} S_{in}^{2}(\theta) - \sum_{i,p \in A} S_{ip}^{2}(\theta)$, defined over a set $A$.\looseness-1 
\end{theorem}
\textbf{Triplet Loss} :  Here we provide the proof of non-submodularity for the Triplet loss discussed in \cref{thm:triplet}.
\begin{proof}
    We first show that the Triplet loss is not necessarily submodular. The reason for this is the Triplet loss is of the form: $\sum_{n \in \mathcal V} \sum_{i, p \in A} D_{in} - \sum_{i, p \in A} D_{ip}$. Note that this is actually supermodular since $-\sum_{i, p \in A} D_{ip}$ is submodular and $\sum_{i, p, n \in A} D_{ip}$ is submodular. As a result, the Triplet loss is \textbf{not necessarily submodular}. 
\end{proof}

\textbf{Submod-Triplet} : Here we provide the proof for the Submod-Triplet loss in \cref{thm:triplet}.
\begin{proof}
    Submodular Triplet loss (Submod-Triplet) is exactly the same as Graph-Cut where we use $\lambda = 1$ and the similarity as the squared similarity function. Thus, this function is \textbf{submodular} in nature.
\end{proof}

\subsubsection{Soft-Nearest Neighbor (SNN) loss and Submod-SNN loss}
\label{app:proof_submod_snn}
\begin{theorem}
\label{thm:snn}
    The Soft-Nearest Neighbor (SNN) loss $L(\theta)$, depicted in row 3 of Table \ref{tab:comb_overview} is not submodular in its original form while the slightly modified form, Submod-SNN loss $L(\theta) = \sum_{k=1}^{|C|} \frac{1}{|A_k|} \sum_{i \in A_{k}} [\log \sum_{j \in A_{k}} \exp(D_{ij}(\theta)) + \log\sum_{j \in \mathcal{V} \setminus A_{k}} \exp(S_{ij}(\theta))]$ is an instance of \score\ ($S_f$ version) with the submodular function $f(A_k; \theta) = \sum_{i \in A_{k}} [\log \sum_{j \in A_{k}} \exp(D_{ij}(\theta)) + \log\sum_{j \in \mathcal{V} \setminus A_{k}} \exp(S_{ij}(\theta))]$.\looseness-1 
\end{theorem}
\textbf{SNN Loss}: Here we provide the proof of non-submodularity for the SNN loss discussed in \cref{thm:snn}.
\begin{proof}
    From the set representation of the SNN loss we can describe the objective $L(\theta)$ as in Equation \ref{eq:snn} . This objective function can be split into two distinct terms labelled as \textit{Term 1} and \textit{Term 2} in the equation above.\\
\begin{equation}
    L(\theta) =\sum_{k = 1}^{|C|} -\underbrace{\sum_{i \in A_{k}} [\log \sum_{j \in A_{k}} \exp(S_{ij}(\theta))}_\text{Term 1} - \underbrace{\log\sum_{j \in \mathcal{V} \setminus A_{k}} \exp(S_{ij}(\theta))]}_\text{Term 2}
\label{eq:snn}
\end{equation}
We prove the objective to be submodular by considering two popular assumptions : \\
(1) The sum of submodular function over a set of classes $A_i$, $i \in C$, the resultant is submodular in nature. \\
(2) The concave over a modular function is submodular in nature.\\
To prove that $L(\theta, A_k)$ is submodular in nature it is enough to show the individual terms (Term 1 and 2) to be submodular. Note that the sum of submodular functions is submodular in nature.
Considering $F(A) = \sum_{j \in A_k} S_j$ for any given $i \in A_k$, we see that $\log \sum_{j \in A_{k}} \exp(D_{j}(\theta))$ to be modular as it is a sum over terms $\exp(D_{j}(\theta))$. \\
We also know from assumption (2) \cite{fujishige}, that the concave over a modular function is \textit{submodular} in nature, $\log$ being a concave function. 
Thus, $log \sum_{j \in A_k} exp(S_{j})$ is submodular function for a given $i \in A_k$.
Unfortunately, the negative sum over a submodular function cannot be guaranteed to be submodular in nature. This renders SNN to be \textbf{non-submodular} in nature.
\end{proof}

\textbf{Submod-SNN Loss} : Here we provide the proof of submodularity for the submodular function $f(A_k; \theta)$ of the Submod-SNN loss discussed in \cref{thm:snn}.
The variation of SNN loss described in Table \ref{tab:comb_overview} can be represented as an instance of $f(A_k; \theta)$ as shown in Equation \ref{eq:submod_snn}. Similar to the set notation of SNN loss we can split the equation into two terms, referred to as \textit{Term 1} and \textit{Term 2} in the equation above. \\
\begin{equation}
    f(A_k; \theta) = \sum_{i \in A_{k}} [\underbrace{\log \sum_{j \in A_{k}} \exp(D_{ij}(\theta))}_\text{Term 1} + \underbrace{\log\sum_{j \in \mathcal{V} \setminus A_{k}} \exp(S_{ij}(\theta))}_\text{Term 2}]
    \label{eq:submod_snn}
\end{equation}
\begin{proof}
Considering $F(A) = \sum_{j \in A_k} S_j$ for any given $i \in A_k$, we prove $\log \sum_{j \in A_{k}} \exp(D_{j}(\theta))$ to be modular, similar to the case of SNN loss.
Further, using assumption (2) mentioned above we prove that the log (a concave function) over a modular function is submodular in nature. Finally, the sum of submodular functions over a set of classes $A_k$ is submodular according to assumption (1). Thus the term 1, $\sum_{i \in A_{k}} \log \sum_{j \in A_{k}} \exp(D_{ij}(\theta))$ in the equation of Submod-SNN is proved to be submodular in nature.

The term 2 of the equation represents the total correlation function of Graph-Cut ($L_{C_f}(\theta)$). Since graph-cut function has already been proven to be submodular in \citep{fujishige, alt_gen_submod} we prove that term 2 is submodular.

Finally, since the sum of submodular functions is submodular in nature, the sum over term 1 and term 2 which constitutes $L(\theta)$ can also be proved to be \textbf{submodular}. 
\end{proof}

\subsubsection{N-Pairs Loss and Orthogonal Projection Loss (OPL)}
\label{app:proof_submod_npair_opl}
In Table \ref{tab:comb_overview} both N-pairs loss and OPL has been identified to be submodular in nature. In this section we provide proofs of \cref{thm:npairs} and \cref{thm:opl} to show they are submodular in nature.

\textbf{N-pairs Loss} : The N-pairs loss $L(\theta)$ can be represented in set notation as described in Equation \ref{eq:npairs} and has been discussed to be submodular according to \cref{thm:npairs}.
\begin{theorem}
\label{thm:npairs}
    The N-pairs loss $L(\theta) = \sum_{k=1}^{|C|}\frac{-1}{|A_k|}[\sum_{i,j \in A_{k}} S_{ij}(\theta)] + \frac{1}{|A_k|}[\sum_{i \in A_{k}} log(\sum_{j \in \mathcal{V}} S_{ij}(\theta) - 1)]$ is an instance of \score\ ($S_f$ version) with the submodular function $f(A, \theta) = -\sum_{i,j \in A} S_{ij}(\theta) + \sum_{i \in A} log(\sum_{j \in \mathcal{V}} S_{ij}(\theta) - 1)$, defined over set $A$.\looseness-1
\end{theorem}
\begin{proof}
    Similar to SNN loss, we can split the equation for $f(A_k; \theta)$ into two distinct terms.
    \begin{equation}
        f(A_k; \theta) = \underbrace{-[\sum_{i,j \in A_{k}} S_{ij}(\theta)}_\text{Term 1} + \underbrace{\sum_{i \in A_{k}} log(\sum_{j \in \mathcal{V}} S_{ij}(\theta) - 1)]}_\text{Term 2}
        \label{eq:npairs}
    \end{equation}
    The first term (Term 1) in N-pairs is a negative sum over similarities, which is submodular in nature \cite{fujishige}. 
    The second term (Term 2) is a log over $\sum_{j \in \mathcal{V}} S_{ij}(\theta) - 1$, which is a constant term for every training iteration as it encompases the whole ground set $\mathcal{V}$. The sum of Term 1 and Term 2 over a set $A_k$ is thus \textbf{submodular} in nature.
\end{proof}

\textbf{OPL} : The Orthogonal Projection loss can be represented as Equation \ref{eq:opl} in its original form and discussed to be submodular according to \cref{thm:opl} in the main paper. 
\begin{theorem}
\label{thm:opl}
    The Orthogonal Projection loss (OPL) $L(\theta) = \sum_{k=1}^{|C|} \frac{1}{|A_k|} ( 1 - \sum_{i,j \in A_{k}} S_{ij}(\theta)) + \frac{1}{|A_k|}\sum_{i \in A_{k}} \sum_{j \in \mathcal{V} \setminus A_{k}} S_{ij}(\theta)$ is an instance of \score ($S_f$ version). OPL largely represents the Graph-Cut (GC) function which is also an instance of \score\ with the submodular function $f(A, \theta) = ( 1 - \sum_{i,j \in A} S_{ij}(\theta)) + \sum_{i \in A} \sum_{j \in \mathcal{V} \setminus A} S_{ij}(\theta)$, defined over a set $A$ with $N_f(A_k) = |A_k|$.\looseness-1
\end{theorem}
\begin{proof}
    Similar to above objectives we split the equation for $f(A_k; \theta)$ into two distinct terms and individually prove them to be submodular in nature.
    \begin{equation}
        f(A_k; \theta) = ( 1 \underbrace{- \sum_{i,j \in A_{k}} S_{ij}(\theta))}_\text{Term 1} + \underbrace{(\sum_{i \in A_{k}} \sum_{j \in \mathcal{V} \setminus A_{k}} S_{ij}(\theta))}_\text{Term 2}
        \label{eq:opl}
    \end{equation}
    The Term 1 represents a negative sum over similarities in set $A_k$ and is thus submodular in nature. The Term 2 is exactly $L_{C_f}$ of Graph-Cut (GC) with $\lambda = 1$ and is also submodular in nature. Since the sum of two submodular functions is also submodular, the underlying function $f(A_k; \theta)$ in $L(\theta, A_k)$ as in \cref{eq:opl} is also \textbf{submodular}.
\end{proof}

\subsubsection{Supervised Contrastive (SupCon) Loss}
\label{app:proof_submod_supcon}
\begin{theorem}
\label{thm:supcon}
    The SupCon loss $L(\theta) = \sum_{k = 1}^{|C|} [\frac{-1}{|A_{k}|} \sum_{i,j \in A_{k}} S_{ij}(\theta) ] + \sum_{i \in A_{k}}\frac{1}{|A_{k}|}\log(\sum_{j \in V}exp(S_{ij}(\theta)) - 1)$ depicted in row 4 of Table \ref{tab:comb_overview} is an instance of \score\ ($S_f$ version) with the submodular function $f(A_k, \theta) = -[\sum_{i,j \in A_{k}} S_{ij}(\theta) ] + \sum_{i \in A_{k}} [ log (\sum_{j \in \mathcal{V} \setminus A_{k}} \exp(S_{ij}(\theta)) - 1)]$, defined over a set $A$ and normalization factor $N_f(A_k) = |A_k|$.\looseness-1 
\end{theorem}
The combinatorial formulation of SupCon as in Equation \ref{eq:set_supcon} can be defined as a sum over the set-function $L(\theta, A_k)$ as described in \cref{thm:supcon} of the main paper. 
\begin{equation}
    f(A_k; \theta) = \underbrace{-\sum_{i,j \in A_{k}} S_{ij}(\theta)}_\text{Term 1} + \underbrace{\sum_{i \in A_{k}}\log(\sum_{j \in V}exp(S_{ij}(\theta)) - 1)}_\text{Term 2}
    \label{eq:set_supcon}
\end{equation}

\begin{proof}
The Term 1 of SupCon is a negative sum over similarities of set $A_k$ and is thus submodular.
The Term 2 contains a sum of the exponent of similarities ($\sum_{j \in V}exp(S_{ij}(\theta)) - 1$) which is a modular term as the sum is computed over the complete ground set $\mathcal{V}$. The logarithm over this term constituting the complete inter-class term represents a concave over modular function which is submodular in nature. 
Thus, the underlying function $f(A_k; \theta)$ for the SupCon loss $L(\theta)$ represented in \cref{thm:supcon} is \textbf{submodular} in nature.\looseness-1
\end{proof}

\subsubsection{SupCon-Var Loss}
\begin{theorem}
\label{thm:supcon_var}
    The SupCon-Var loss $L(\theta) = \sum_{k=1}^{|C|} \frac{-1}{|A_k|}[\sum_{i,j \in A_{k}} S_{ij}(\theta) ] + \frac{1}{|A_k|}\sum_{i \in A_{k}} [ log (\sum_{j \in \mathcal{V} \setminus A_{k}} \exp(S_{ij}(\theta)))]$ is an instance of \score\ ($S_f$ version) with the submodular function $f(A_k, \theta) = -[\sum_{i,j \in A_k} S_{ij}(\theta) ] + \sum_{i \in A_k} [ log (\sum_{j \in \mathcal{V} \setminus A_k} \exp(S_{ij}(\theta)))]$, defined over a set $A_k$, for $k \in [1, C]$ and normalization factor $N_f(A_k) = |A_k|$.\looseness-1 
\end{theorem}
The submodular variant of SupCon (SupCon-Var) as shown in Equation \ref{eq:supcon} can be split into two terms indicated as Term 1 and Term 2.
\begin{equation}
    f(A_k; \theta) = \underbrace{-[\sum_{i,j \in A_{k}} S_{ij}(\theta) ]}_\text{Term 1} + \underbrace{\sum_{i \in A_{k}} [ log (\sum_{j \in \mathcal{V} \setminus A_{k}} \exp(S_{ij}(\theta)) )}_\text{Term 2} ]
    \label{eq:supcon}
\end{equation}
\begin{proof}
The Term 1 of the function $f(A_k; \theta)$ in SupCon-Var is a negative sum over similarities of set $A_k$ and is thus submodular.
The Term 2 of the equation is also submodular as it is a concave over the modular term $\sum_{j \in \mathcal{V} \setminus A_{k}} \exp(S_{ij}(\theta)$, with $\log$ being a concave function.
Thus, the $L_{S_f}$ form of SupCon-Var is also \textbf{submodular} represented as the sum of two submodular functions is submodular in nature.
\end{proof}

\end{document}